# Using Local Alignments for Relation Recognition

**Sophia Katrenko**                                        S.Katrenko@uva.nl
**Pieter Adriaans**                                        P.W.Adriaans@uva.nl
**Maarten van Someren**                                    M.W.vanSomeren@uva.nl
*Informatics Institute, University of Amsterdam*
*Science Park 107, 1098XG Amsterdam, the Netherlands*

## Abstract

This paper discusses the problem of marrying structural similarity with semantic relatedness for Information Extraction from text. Aiming at accurate recognition of relations, we introduce local alignment kernels and explore various possibilities of using them for this task. We give a definition of a local alignment (LA) kernel based on the Smith-Waterman score as a sequence similarity measure and proceed with a range of possibilities for computing similarity between elements of sequences. We show how distributional similarity measures obtained from unlabeled data can be incorporated into the learning task as semantic knowledge. Our experiments suggest that the LA kernel yields promising results on various biomedical corpora outperforming two baselines by a large margin. Additional series of experiments have been conducted on the data sets of seven general relation types, where the performance of the LA kernel is comparable to the current state-of-the-art results.

## 1. Introduction

Despite the fact that much work has been done on automatic relation extraction (or recognition) in the past few decades, it remains a popular research topic. The main reason for the keen interest in relation recognition lies in its utility. Once concepts and semantic relations are identified, they can be used for a variety of applications such as question answering (QA), ontology construction, hypothesis generation and others.

In ontology construction, the relation that is studied most is the IS-A relation (or HYPERNYMY), which organizes concepts in a taxonomy (Snow, Jurafsky, & Ng, 2006). In information retrieval, semantic relations are used in two ways, to refine queries before actual retrieval, or to manipulate the output that is returned by a search engine (e.g. identifying whether a fragment of text contains a given relation or not). The most widely used relations for query expansion are HYPERNYMY (or broader terms from a thesaurus) and SYNONYMY. Semantic relations can also be useful at different stages of question answering. They have to be taken into account when identifying the type of a question and they have to be considered at actual answer extraction time (van der Plas, 2008). Yet another application of relations is constructing a new scientific hypothesis given the evidence found in text. This type of knowledge discovery from text is often based on co-occurrence analysis and, in many cases, was corroborated via experiments in laboratories (Swanson & Smalheiser, 1999).

Another reason why extraction of semantic relations is of interest lies in the diversity of relations. Different relations need different extraction methods. Many existing information extraction systems were originally designed to work for generic data (Grishman & Sundheim, 1996), but it became evident that domain knowledge is often necessary for successful extraction. For instance, relation extraction in the biomedical domain would require an accurate recognition of named entities such as gene names (Clegg, 2008), and in the area of food it needs information on relevant named entities such as toxic substances.

Also for generic relations syntactic information is often not sufficient. Consider, for instance, the following sentences (with the arguments of the relations written in italics):

(1) Mary looked back and whispered: "I know every *tree* in this *forest*, every scent". (PART-WHOLE relation)

(2) A person infected with a particular *flu virus* strain develops antibodies against that virus. (CAUSE-EFFECT relation)

(3) The *apples* are in the *basket*. (CONTENT-CONTAINER relation)

All these sentences exemplify binary relations, namely PART-WHOLE (*tree* is part of a *forest*), CAUSE-EFFECT (*virus* causes *flu*) and CONTENT-CONTAINER (*apples* are contained in *basket*). We can easily notice that the syntactic context in (1) and (3) is the same, namely, the arguments in both cases are connected to each other by the preposition 'in'. However, this context is highly ambiguous because even though it allows us to reduce the number of potential semantic relations, it is still not sufficient to be able to discriminate between PART - WHOLE and CONTENT - CONTAINER relation. In other words, world knowledge about 'trees', 'forests', 'apples' and 'baskets' is necessary to classify relations correctly. The situation changes even more drastically if we consider example (2). Here, there is no explicit indication for CAUSATION. Nevertheless, by knowing what 'a flu' and 'a virus' is, we are able to infer that CAUSE - EFFECT relation holds.

The examples in (1), (2) and (3) highlight several difficulties that characterize semantic relation extraction. Generic relations very often occur in nominal complexes such as 'flu virus' in (2) and lack of sentential context boosts such approaches as paraphrasing (Nakov, 2008). However, even for noun compounds one has to combine world knowledge with the compound's context to arrive at the correct interpretation.

Computational approaches to the relation recognition problem often rely on a two-step procedure. First, the relation arguments are identified. Depending on the relation at hand, this step often involves named entity recognition of the arguments of the relations. The second step is to check whether the relation holds. If relation arguments are provided (e.g., 'basket' and 'apples' in (3)), relation extraction is reduced to the second step. Previous work on relation extraction suggests that in this case the accuracy of relation recognition is much higher than in the case when they have to be discovered automatically (Bunescu et al., 2005). Furthermore, most existing solutions to relation extraction (including work presented in this paper) focus on relation examples that occur within a single sentence and do not consider discourse (McDonald, 2005). Recognizing relations from a wider scope is an interesting enterprise, but it would require to take into account anaphora resolution and other types of linguistic analysis.

Approaches to relation extraction that are based on hand-written patterns are time-consuming and in many cases need an expert to formulate and test the patterns. Although patterns are often precise, they usually produce poor recall (Thomas et al., 2000). In general, hand-written patterns can be of two types. The first type is sequential and based on frequently occurring sequences of words in a sentence. Hand-written sequential patterns were initially used for extraction of Hypernymy (Hearst, 1992), with several attempts to extend them to other relations. The second type of patterns (Khoo, Chan, & Niu, 2000) take the syntactic structure of a sentence into account. The dependency structure of a sentence can usually be represented as a tree and the patterns then become subtrees. Such patterns are sometimes referred to as graphical patterns. To identify examples of the Cause-Effect relation, Khoo et al. (2000) applied this type of patterns to texts in the medical domain. This study showed that graphical patterns are sensitive to the errors made by the parsers, do not cover all examples in the test data and extract many spurious instances.

An alternative to using hand-written patterns is supervised Machine Learning. Then, relations are labeled and used to train a classifier that can recognize these relations in new texts. One approach is to learn generalized extraction patterns where patterns are expressed as characters, words or syntactic categories of words. Other approaches involve clustering based on co-occurrence (Davidov & Rappoport, 2008). In recent years kernel-based methods have become popular because they can handle high-dimensional problems (Zelenko et al., 2003; Bunescu & Mooney, 2006; Airola et al., 2008). These methods transform text fragments, complete sentences or segments around named entitites or verbs, to vectors, and apply Support Vector Machines to classify new fragments.

Some Machine Learning methods use prior knowledge that is given to the system in addition to labeled examples (Schölkopf, 1997, p. 17). The use of prior knowledge is often motivated by, for example, poor quality of data and data sparseness. Prior knowledge can be used in many ways, from changing the representation of existing training examples to adding more examples from unlabelled data. For NLP tasks, prior knowledge exists in the form of manually (or automatically) constructed ontologies or large collections of unannotated data. These enrich the textual data and thereby improve the recognition of relations (Sekimizu, Park, & Tsujii, 1998; Tribble & Fahlman, 2007). Recently, Zhang et al. (2008) showed that semantic correlation of words can be learned from unlabeled text collections, transferred among documents and used further to improve document classification. In general, while use of large collections of text allows us to derive almost any information needed, it is done with varying accuracy. In contrast, existing resources created by humans can provide very precise information, but it is less likely that they will cover all possible areas of interest.

In this paper, as in the work of Bunescu and Mooney (2006), we use the syntactic structure of sentences, in particular, dependency paths. This stems from the observation that linguistic units are organized in complex structures and understanding how words or word senses relate to each other often requires contextual information. Relation extraction is viewed as a supervised classification problem. A training set consists of examples of a given relation and the goal is to construct a model that can be applied to a new, unseen data set, to recognize all instances of the given relation in this new data set. For recognition of relations we use a kernel-based classifier that is applied to dependency paths. However, instead of a vector-based kernel we directly use similarity between dependency paths and show how information from existing ontologies or large text corpora can be employed.

The paper is organized as follows. We start by reviewing existing kernel methods that work on sequences (Section 2). In Section 3, we give the definition of a local alignment kernel based on the Smith-Waterman measure. We proceed by discussing how it can be used in the context of natural language processing (NLP) tasks, and particularly for extracting relations from text (Section 3.2). Once the method is described, we report on two types of the data sets (biomedical and generic) used in the experiments (Section 4) and elaborate on our experiments (Sections 5 and 6). Section 7 discusses our findings in more detail. Section 8 concludes the paper by discussing possible future directions.

## 2. Kernel Methods

The past years have witnessed a boost of interest in kernel methods, their theoretical analysis and practical applications in various fields (Burges, 1998; Shawe-Taylor & Christianini, 2000). The idea of having a method that works with different structures and representations, starting from the simplest representation using a limited number of attributes to complex structures such as trees, seems indeed very attractive.

Before we define a kernel function, recall the standard setting for supervised classification. For a training set $S$ of $n$ objects (instances) $(\mathbf{x_1}, y_1), \ldots, (\mathbf{x_n}, y_n)$ where $\mathbf{x_1}, \ldots, \mathbf{x_n} \in \mathcal{X}$ are input examples in the input space $\mathcal{X}$ with their corresponding labels $y_1, \ldots, y_n \in \{0,1\}$, the goal is to infer a function $h : \mathcal{X} \to \{0, 1\}$ such that it approximates a target function $t$. However, $h$ can still err on the data which has to be reflected in a loss function, $l(h(\mathbf{x_i}), y_i)$. Several loss functions have been proposed in the literature so far, the best known of which is the zero-one loss. This loss is a function that outputs 1 each time a method errs on a data point $(h(\mathbf{x_i}) \neq y_i)$, and 0 otherwise.

The key idea of kernel methods lies in the implicit mapping of objects to a high-dimensional space (by using some mapping function $\phi$) and considering their inner product (similarity) $k(\mathbf{x_i}, \mathbf{x_j}) = <\phi(\mathbf{x_i}), \phi(\mathbf{x_j})>$, rather than representing them explicitly. Functions that can be used in kernel methods have to be symmetric and positive semi-definite, whereby positive semi-definiteness is defined by $\sum_{i=1}^{n} \sum_{j=1}^{n} c_i c_j k(\mathbf{x_i}, \mathbf{x_j}) \geq \mathbf{0}$ for any $n > 0$, any objects $\mathbf{x_1}, \ldots, \mathbf{x_n} \in \mathcal{X}$, and any choice of real numbers $c_1, \ldots, c_n \in \mathbb{R}$. If a function is not positive semi-definite, the algorithm may not find the global optimal solution. If the requirements w.r.t. symmetry and positive semi-definiteness are met, a kernel is called *valid*.

Using the idea of a kernel mapping, Cortes and Vapnik (1995) introduced support vector machines (SVM) as a method which seeks the linear separation between two classes of the input points by a function $f(x)$ such that $f(x) = \mathbf{w}^T \phi(\mathbf{x}) + b$, $\mathbf{w}^T \in \mathbb{R}^p$, $b \in \mathbb{R}$ and $h(x) = sign(f(x))$. Here, $\mathbf{w}^T$ stands for the slope of the linear function and $b$ for its offset. Often, there can exist several functions that separate data well, but not all of them are equally good. A hyperplane that separates mapped examples with the largest possible margin would be the best option (Vapnik, 1982).

SVMs solve the following optimization problem:

$$\underset{f(x)=\mathbf{w}^T x + b}{\operatorname{argmin}} \frac{1}{2} \parallel \mathbf{w} \parallel^2 + C \sum_{i=1}^{n} l(h(x_i), y_i) \tag{4}$$

In Equation 4, the first part of the equation corresponds to the margin maximization (by minimizing $\frac{1}{2} \parallel \mathbf{w} \parallel^2$), while the second takes into account the error on the training set which has to be minimized (where $C$ is a penalty term). The hyperplane that is found may correspond to a non-linear boundary in the original input space. There exist a number of standard kernels such as the linear kernel, the Gaussian kernel and others. Information about the data or the problem can motivate the choice of a particular kernel. It has been shown by Haussler (1999) that a complex kernel (referred to as a *convolution kernel*) can be defined using simpler kernels.

Other forms of machine learning representations for using prior knowledge were defined along with the methods for exploiting it. Inductive logic programming offers one possible solution to use it explicitly, in the form of additional Horn clauses (Camacho, 1994). In the Bayesian learning paradigm information on the hypothesis without seeing any data is encoded in a Bayesian prior (Mitchell, 1997) or in a higher level distribution in a hierarchical Bayesian setting. It is less obvious though how to represent and use prior knowledge in other learning frameworks. In the case of SVMs, there are three possible ways of incorporating prior knowledge (Lauer & Bloch, 2008). These are named sampling methods (prior knowledge is used here to generate new data), kernel methods (prior knowledge is incorporated in the kernel function by, for instance, creating a new kernel), and optimization methods (prior knowledge is used to reformulate the optimization problem by, for example, adding additional constraints). The choice of a kernel can be based on general statistical properties of the domain, but an attractive possibility is to incorporate explicit domain knowledge into the kernel. This can improve a kernel by "smoothing" the space: instances that are more similar have a higher probability of belonging to the same class than with a kernel without prior knowledge.

In what follows, we review a number of kernels on strings that have been proposed in the research community over the past years. A very natural domain to look for them is the biomedical field where many problems can be formulated as string classification (protein classification and amino acid sequences, to name a few). Sequence representation is, however, not only applicable to the biomedical area, but can also be considered for many natural language processing tasks. After introducing kernels that have been used in biomedicine, we move to the NLP domain and present recent work on relation extraction employing kernel methods.

## 2.1 The Spectrum Kernel

Leslie, Eskin, and Noble (2002) proposed a discriminative approach to protein classification. For any sequence $\mathbf{x} \in \mathcal{X}$, the authors define the $m$-spectrum as the set $S$ of all contiguous subsequences of $\mathbf{x}$ whose length is equal to $m$. All possible $m$-long subsequences $q \in S$ are indexed by the frequency of their occurrence ($\phi_q(\mathbf{x})$). Consequently, a feature map for a sequence $\mathbf{x}$ and alphabet $A$ equals $\Phi_m(x) = (\phi_q(\mathbf{x}))_{q \in A^m}$. The spectrum kernel for two sequences $\mathbf{x}$ and $\mathbf{y}$ is defined as the inner product between the corresponding feature maps: $k_S(x, y) = <\Phi_m(\mathbf{x}), \Phi_m(\mathbf{y})>$.

Now, even assuming contiguous subsequences for small $m$, the feature space to consider is very large. The authors propose to detect all subsequences of length $m$ by using a suffix tree method which guarantees fast computation of the kernel matrix. The spectrum kernel was tested on the task of protein homology detection, where the best results were achieved by setting $m$ to a relatively small number (3). The novelty of Leslie et al.'s (2002) method lies in its generality and its low computational complexity.

## 2.2 Mismatch Kernels

The mismatch kernel that was introduced later by Leslie et al. (2004) is essentially an extension of the latter. An obvious limitation of the spectrum kernel is that all considered subsequences are contiguous and should match exactly. In the mismatch kernel the contiguity is preserved while the match criterion is changed. In other words, instead of looking for all possible subsequences of length $m$ for a given subsequence, one is searching for all possible subsequences of length $m$ allowing up to $r$ mismatches. Such a comparison will result in a larger subset of subsequences, but the kernels defined in this way can still be calculated rather fast. The kernel is formulated similarly to the spectrum kernel and the only major difference is in computing the feature map for all sequences. More precisely, a feature map for a sequence $\mathbf{x}$ is defined as $\Phi_{m,r}(\mathbf{x}) = \sum_{q \in S} \Phi_{m,r}(q)$ where $\Phi_{m,r}(q) = (\phi_\beta(q))_{\beta \in A^m}$. $\phi_\beta(q)$ is binary and indicates whether sequence $\beta$ belongs to the set of $m$-length sequences that differ from $q$ at most in $r$ elements (1) or it does not (0). It is clear that if $r$ is set to 0, the mismatch kernel is reduced to the spectrum kernel. The complexity of the mismatch kernel computation is linear with respect to the sum of the sequence lengths.

The authors also show that the mismatch kernel not only yields state-of-the-art performance on a protein classification task but also outputs subsequences that are informative from a biological point of view.

## 2.3 Kernel Methods and NLP

One of the merits of kernel methods is the possibility of designing kernels for different structures, such as strings or trees. In the NLP field (and in relation extraction, in particular) most work roughly falls into two categories. In the first, kernels are defined over the plain text using sequences of words. The second uses linguistic structures such as dependency paths or trees or the output of shallow parsing. In this short review we do not take a chronological perspective but rather start with the methods that are based on sequences and proceed with the approaches that make use of syntactic information.

In the same year in which the spectrum kernel was designed, Lodhi et al. (2002) introduced string subsequence kernels that provide flexible means to work with text data. In particular, subsequences are not necessarily contiguous and are weighted according to their length (using a decay factor $\lambda$). The length of the subsequences is fixed in advance. The authors claim that even without the use of any linguistic information their kernels are able to capture semantic information. This is reflected in the better performance on the text classification task compared to the bag-of-words approach. While Lodhi et al.'s (2002) kernel works on sequences of characters, a kernel proposed by Cancedda et al. (2003) is applied to word sequences. String kernels can be also extended to syllable kernels which proved to do well on text categorization (Saunders, Tschach, & Shawe-Taylor, 2002).

Because all these kernels can be defined recursively, their computation is efficient. For instance, the time complexity of Lodhi et al.'s (2002) kernel is $\mathrm{O}(n|s||t|)$, where $n$ is the length of the subsequence, and $t$ and $s$ are documents.

### 2.3.1 Subsequence Kernels

In the recognition of binary relations, the most natural way is to consider words located around and between relation arguments. This approach was taken by Bunescu and Mooney (2005b) whose choice of sequences was motivated by textual patterns found in corpora. For instance, they observed that some relations are expressed by 'subject-verb-object' constructions while others are part of the noun and prepositional phrases. As a result, three types of sequences were considered: fore-between (words before and between two named entities), between (words only between two entities) and between-after (words between and after two entities). The length of sequences is restricted. To handle data sparseness, the authors generalize over existing sequences using PoS tags, entity types and WordNet synsets. A generalized subsequence kernel is recursively defined as the number of weighted sparse subsequences that two sequences share. In the absence of syntactic information, an assumption is made that long subsequences are not likely to represent positive examples and as such are penalized. This subsequence kernel is computed for all three types of sequences and the resulting relation kernel is defined as a sum over the three subkernels. Experimental results on a biomedical corpus are encouraging, showing that the relation kernel performs better than manually written patterns and an approach based on longest common subsequences.

A method proposed by Giuliano et al. (2006) was largely inspired by the work of Bunescu and Mooney (2005b). However, instead of looking for subsequences in three types of sequences, the authors treat them as a bag-of-words and define what is called a global kernel as follows. First, each sequence type (pattern) $P$ is represented by a vector whose elements are counts of how many times each token was used in $P$. A local kernel is defined similarly but only using words surrounding named entities (left and right context). A final shallow linguistic kernel is defined as the combination of the global and the local kernels. Experiments on biomedical corpora suggest that this kernel outperforms the subsequence kernel by Bunescu and Mooney.

### 2.3.2 Distributional Kernels

Recently, Ó Séaghdha and Copestake (2008) introduced distributional kernels on co- occurrence probability distributions. The co-occurrence statistics they use are in the form of either syntactic relations or $n$-grams. They show that it is possible to derive kernels from such distances as Jensen-Shannon divergence (JSD) or Euclidean distance ($L_2$) (Lee, 1999). JSD is a smoothed version of the Kullback-Leibler divergence, an information-theoretic measure of the dissimilarity between two probability distributions. The main motivation behind this approach lies in the fact that distributional similarity measures proved to be useful for NLP tasks. To extract co-occurrence information, the authors use two corpora, the British National Corpus (BNC) and the Web 1T 5-Gram Corpus (which contains 5-grams with their observed frequency counts and was collected from the Web). Distributional kernels proved to be successful for a number of tasks such as compound interpretation, relation extraction and verb classification. On all of them, the JSD kernel clearly outperforms Gaussian and linear kernels. Moreover, estimating distributional similarity on the BNC corpus yields performance similar to the results obtained on the Web 1T 5-Gram Corpus. This is an interesting finding because the BNC corpus was used to estimate similarity from syntactic relations whereas the latter corpus contains $n$-grams only. Most importantly, the method of Ó Séaghdha and Copestake provides empirical support for the claim that using distributional similarity is beneficial for relation extraction.

### 2.3.3 Kernels for Syntactic Structures

Kernels defined for unpreprocessed text data seem attractive because they can be applied directly to text from any language. However, as general as they are, they can lose precision when compared to the methods that use syntactic analysis. Re-ranking parsing trees (Collins & Duffy, 2001) was one of the first applications of kernel methods to NLP problems. To accomplish this goal, the authors rely on the subtrees that a pair of trees have in common. Later on, Moschitti (2006) explored convolution kernels on dependency and constituency structures to do semantic role labeling and question classification. This work introduces a novel kernel which is called a partial tree kernel (PT). It is essentially built on two kernels proposed before, the subtree kernel (ST) that contains all descendant nodes from a target root (including leaves) and the subset tree kernel (SST) that is more flexible and allows internal subtrees which do not necessarily encompass leaves. A partial tree is a generalization of a subset tree whereby partial structures of a grammar are allowed (i.e., parts of the production rules such as [VP [V]] form a valid PT). Moschitti demonstrated that PTs obtain better performance on dependency structures than SSTs, but the latter yield better results on constituent trees.

### 2.3.4 Kernel on Shallow Parsing Output

Zelenko et al. (2003) use shallow parsing and designed kernels to extract relations from text. In contrast to full parsing, shallow parsing produces partial interpretations of sentences. Each node in such a tree is enriched with information on roles (that correspond to the arguments of a relation). The similarity of two trees is determined by the similarity of their nodes. Depending on how similarity is computed, Zelenko et al. define two types of kernels, contiguous subtree kernels and sparse kernels. Both types were tested on two types of relations, 'person-affiliation' and 'organization-location' exhibiting good performance. In particular, sparse kernels outperform contiguous subtree kernels leading to the conclusion that partial matching is important when dealing with typically sparse natural language data. However, the computation of the sparse kernel takes $O(mn^3)$ time (where $m$ and $n$ are the number of children of two relation examples, i.e. shallow trees, under consideration, $m \geq n$), while the algorithm for the contiguous subtree kernel runs in time $O(mn)$.

### 2.3.5 Shortest Path Kernel

Bunescu and Mooney's (2005a) shortest path kernel represents yet another approach for relation extraction that is kernel-based and relies on information found in dependency trees. A main assumption here is that not the entire dependency structure is relevant, and one can focus on the path that is connecting two relation arguments instead. The more similar these paths are, the more likely two relation examples belong to the same category. In spirit with their previous work, Bunescu and Mooney seek generalizations over existing paths by adding information sources like part of speech (PoS) categories or named entity types.

The shortest path between relation arguments is extracted and a kernel between two sequences (paths) $\mathbf{x} = \{x_1, \ldots, x_n\}$ and $\mathbf{x}' = \{x_1', \ldots, x_m'\}$ is computed as follows:

$$k_B(\mathbf{x}, \mathbf{x}') = \begin{cases} 0 & m \neq n \\ \prod_{i=1}^{n} f(x_i, x_i') & m = n \end{cases} \tag{5}$$

In Equation 5, $f(x_i, x_i')$ is the number of features shared by $x_i$ and $x_i'$. Bunescu and Mooney (2005a) use several features such as word (e.g., $protesters$), part of speech tag (e.g., $NNS$), generalized part of speech tag (e.g., $Noun$), and entity type (e.g., $PERSON$) if applicable. In addition, a direction feature ($\rightarrow$ or $\leftarrow$) is employed. Here we reproduce an example from their paper.

**Example 1** *Given two dependency paths that exemplify the relation* `Located` *such as 'his $\rightarrow$ actions $\leftarrow$ in $\leftarrow$ Brcko' and 'his $\rightarrow$ arrival $\leftarrow$ in $\leftarrow$ Beijing', both paths are expanded by additional features as those mentioned above. It is easy to see that comparing path (6) to path (7) gives us a score of 18 ($3 \times 1 \times 1 \times 1 \times 2 \times 1 \times 3 = 18$).*

$$\begin{bmatrix} his \\ PRP \\ PERSON \end{bmatrix} \times [\rightarrow] \times \begin{bmatrix} actions \\ NNS \\ Noun \end{bmatrix} \times [\leftarrow] \times \begin{bmatrix} in \\ IN \end{bmatrix} \times [\leftarrow] \times \begin{bmatrix} Brcko \\ NNP \\ Noun \\ LOCATION \end{bmatrix} \tag{6}$$

$$\begin{bmatrix} his \\ PRP \\ PERSON \end{bmatrix} \times [\rightarrow] \times \begin{bmatrix} arrival \\ NN \\ Noun \end{bmatrix} \times [\leftarrow] \times \begin{bmatrix} in \\ IN \end{bmatrix} \times [\leftarrow] \times \begin{bmatrix} Beijing \\ NNP \\ Noun \\ LOCATION \end{bmatrix} \tag{7}$$

The time complexity of the shortest path kernel is O($n$), where $n$ stands for the length of the dependency path.

Dependency paths are also considered in other recent work on relation recognition (Erkan, Özgür, & Radev, 2007). Here, Erkan et al. (2007) use dependency paths as input and compare them by means of cosine similarity or edit distance. The authors motivate their choice by the need to compare dependency paths of different length. Further, various machine learning methods are used to do classification, including SVM and transductive SVM (TSVM), which is an extension of SVM (Joachims, 1999). In particular, TSVM makes use of labeled and unlabeled data by first classifying the unlabeled examples and then searching for the maximum margin that separates positive and negative instances from both sets. The authors conclude that edit distance performs better than the cosine similarity measure, and that TSVM slightly outperforms SVM.

Airola et al. (2008) propose a graph kernel which makes use of the entire dependency structure. In their work, each sentence is represented by two subgraphs, one of which is built from the dependency analysis, and the other corresponds to the linear structure of the sentence. Further, a kernel is defined on all paths between any two vertices in the graph. The method by Airola et al. (2008) achieves state-of-the-art performance on biomedical data sets, and is further discussed, together with the shortest path kernel and the work by Erkan et al. (2007), in Section 5 on relation extraction in the biomedical domain in this paper.

Finally, kernels can be defined not only on graphs of syntactic structures, but also on graphs of a semantic network. This is illustrated by Ó Séaghdha (2009), who uses graph kernels on the graph built from the HYPONYMY relations in WordNet. Even though no syntactic information is utilized, such kernels proved to perform well on the extraction of various generic relations.

All kernels that we reviewed in this section deal with sequences or trees albeit in different ways. The empirical findings suggest that kernels that allow partial matching usually perform better when compared to methods where similarity is defined on an exact match. To alleviate the problem of exact matching, some researchers suggested generalizing over elements in existing structures (Bunescu & Mooney, 2005a) while others opted for a flexible comparison. In our view, these types of methods can complement each other (Saunders et al., 2002). As flexible as the partial matching methods are, they may suffer from low precision when the penalization of the mismatch is low. The same holds for approaches that use generalization strategies because they may easily overgeneralize. A possible solution would be to combine both, provided that mismatches are penalized well and generalizations are semantically plausible rather than based on part of speech categories. This idea is further explored in the present paper and evaluated on the relation recognition task.

In a nutshell, the goals of this paper are the following: (i) a study of the possibilities of using the local alignment kernel for relation extraction from text, (ii) an exploration of the use of prior knowledge in the alignment kernel and (iii) an extensive evaluation with automatic recognition of two types of relations, biomedical and generic.

## 3. A Local Alignment Kernel

One can note from our short overview of the kernels designed for NLP above that many researchers use partial structures and propose variants such as subsequence kernels (Bunescu & Mooney, 2005b), a partial tree kernel (Moschitti, 2006), or a kernel on shallow parsing output (Zelenko et al., 2003) for relation extraction. In this paper we focus on dependency paths as input and formulate the following requirements for a kernel function:

- it should allow partial matching so that the similarity can be measured for paths of different length

- it should be possible to incorporate prior knowledge

Recall that by prior knowledge we mean information that comes either from larger corpora or from existing resources such as ontologies. For instance, knowing that 'development' is synonymous to 'evolution' in some contexts can help to recognize that two different words are close semantically. Such information is especially useful if the meaning is relevant for detecting relations that may differ in form.

In the following subsection we will define a local alignment kernel that satisfies these requirements and show how to incorporate prior knowledge.

### 3.1 Smith-Waterman Measure and Local Alignments

Our work here is motivated by the recent advances in the biomedical field. It has been shown that it is possible to design valid kernels based on a similarity measure for strings (Saigo, Vert, & Akutsu, 2006). For example, Saigo, Vert, Ueda, and Akutsu (2004) consider the Smith-Waterman (SW) similarity measure (Smith & Waterman, 1981) (see below) to measure the similarity between two sequences of amino acids.

String distance measures can be divided into measures based on terms, edit-distance and Hidden Markov models (HMM) (Cohen, Ravikumar, & Fienberg, 2003). Term-based distances such as measures based on the TF-IDF score, consider a pair of word sequences as two sets of words ignoring their order. In contrast, string edit distances (or string similarity measures) treat entire sequences and compare them using transformation operations, which convert a sequence $\mathbf{x}$ into a sequence $\mathbf{x}'$. Examples of these are the Levenshtein distance, and the Needleman-Wunsch (Needleman & Wunsch, 1970) and Smith-Waterman (Smith & Waterman, 1981) measures. The Levenshtein distance has been used in the natural language processing field as a component in a variety of tasks, including semantic role labeling (Sang et al., 2005), construction of paraphrase corpora (Dolan, Quirk, & Brockett, 2004), evaluation of machine translation output (Leusch, Ueffing, & Ney, 2003), and others. The Smith-Waterman measure is mostly used in the biological domain, there are, however, some applications of a modified Smith-Waterman measure to text data as well (Monge & Elkan, 1996; Cohen et al., 2003). HMM-based measures present probabilistic extensions of edit distances (Smith, Yeganova, & Wilbur, 2003).

Our hypothesis is that string similarity measures are the best basis for a kernel for relation extraction. In this case, the order in which words appear is likely to be relevant and sparse data usually prevents estimation of probabilities (as in the work of Smith et al., 2003). In general, two sequences can be aligned in several possible ways. It is possible to search either for an alignment which spans entire sequences (global alignment), or for an alignment which is based on similar subsequences (local alignment). Both in the case of sequences of amino acids and in relation extraction, local patterns are likely to be the most important factor that determines similarity. Therefore we need a similarity measure that emphasizes local alignments.

Formally, we define a pairwise alignment $\pi$ of at most $L$ elements for two sequences $\mathbf{x} = x_1 x_2 \ldots x_n$ and $\mathbf{x}' = x_1' x_2' \ldots x_m'$, as a pairing $\pi = \{\pi_l(i,j)\}$, $l = 1, \ldots, L$, $1 \leq i \leq n$, $1 \leq j \leq m$, $1 \leq l \leq n$, $1 \leq l \leq m$. In Example 2 (ii), the third element of the first sequence is aligned with the first element of the second one, which is denoted by $\pi_1(3,1)$.

**Example 2** *Given the sequences* $\mathbf{x}$=*abacde and* $\mathbf{x}'$=*ace, two possible alignments (with gaps indicated by '-') are as follows.*

*(i) global alignment*

$$
\begin{array}{cccccc}
a & b & a & c & d & e \\
a & \text{-} & \text{-} & c & \text{-} & e
\end{array}
\qquad \textit{Alignment:} \quad \pi = \{\pi_1(1,1), \pi_2(4,2), \pi_3(6,3)\}
$$

*(ii) local alignment*

$$
\begin{array}{cccccc}
a & b & a & c & d & e \\
\text{-} & \text{-} & a & c & \text{-} & e
\end{array}
\qquad \textit{Alignment:} \quad \pi = \{\pi_1(3,1), \pi_2(4,2), \pi_3(6,3)\}
$$

In this example, the number of gaps inserted in $\mathbf{x}'$ to align it with $\mathbf{x}$ and the number of elements that match is the same in both cases. Yet, both in the biological and in the linguistic context we may prefer alignment (ii), because closely matching substrings, local alignments, are a better indicator for similarity than shared items that are far apart. It is, therefore, better to use a measure that puts less or no weight on gaps before the start or after the end of strings (as in Example 2 (ii)). This can be done using a local alignment mechanism that searches for the most similar subsequences in two sequences. Local alignments are employed when sequences are dissimilar and are of different length, while global alignments are considered when sequences are of roughly the same length. From the measures we have mentioned above, the Smith-Waterman measure is a local alignment measure, and the Needleman-Wunsch measure compares two sequences based on global alignments.

**Definition 1 (Global alignment)** *Given two sequences $\mathbf{x} = x_1 \ldots x_n$ and $\mathbf{x}' = x'_1 \ldots x'_m$, their global alignment is a pair of sequences $\mathbf{y}$ and $\mathbf{y}'$ both of the same length, which are obtained by inserting zero or more gaps before the first element of either $\mathbf{x}$ or $\mathbf{x}'$, and after each element of $\mathbf{x}$ and of $\mathbf{x}'$.*

**Definition 2 (Local alignment)** *Given two sequences $\mathbf{x} = x_1 \ldots x_n$ and $\mathbf{x}' = x'_1 \ldots x'_m$, their local alignment is a pair of subsequences $\alpha$ of $\mathbf{x}$ and $\gamma$ of $\mathbf{x}'$, whose similarity is maximal.*

To clarify what we mean by local and global alignments, we give a definition of both the Smith-Waterman and Needleman-Wunsch measures. Given two sequences $\mathbf{x} = x_1 x_2 \ldots x_n$ and $\mathbf{x}' = x'_1 x'_2 \ldots x'_m$ of length $n$ and $m$ respectively, the Smith-Waterman measure is defined as a similarity score of their best local alignment:

$$\mathrm{sw}(\mathbf{x}, \mathbf{x}') = \max_{\pi \in A(\mathbf{x}, \mathbf{x}')} s(\mathbf{x}, \mathbf{x}', \pi) \tag{8}$$

In the equation above, $s(\mathbf{x}, \mathbf{x}', \pi)$ is a score of a local alignment $\pi$ of sequence $\mathbf{x}$ and $\mathbf{x}'$ and $A$ denotes the set of all possible alignments. The best local alignment can be efficiently found using dynamic programming. To do this, one fills in a matrix $SW$ with partial alignments as follows:

$$\mathop{\mathrm{SW}}_{\substack{1 \le i \le n, \\ 1 \le j \le m}} (i, j) = \max \begin{cases} 0 \\ \mathrm{SW}(i-1, j-1) + d(x_i, x'_j) \\ \mathrm{SW}(i-1, j) - G \\ \mathrm{SW}(i, j-1) - G \end{cases} \tag{9}$$

In Equation 9, $d(x_i, x'_j)$ denotes a substitution score between two elements $x_i$ and $x'_j$ and $G$ stands for a gap penalty. Using this equation it is possible to find partial alignments, that are stored in a matrix in which the cell $(i, j)$ reflects the score for alignment between $x_1 \ldots x_i$

|   | a | b | a | c | d | e |
|---|---|---|---|---|---|---|
|   | 0 | 0 | **0** | 0 | 0 | 0 | 0 |
| a | 0 | 2 | 1 | **2** | 1 | 0 | 0 |
| c | 0 | 1 | 1 | 1 | **4** | **3** | 2 |
| e | 0 | 0 | 0 | 0 | 3 | 3 | **5** |

(a) Smith-Waterman measure

|   | a | b | a | c | d | e |
|---|---|---|---|---|---|---|
|   | **0** | 0 | 0 | 0 | 0 | 0 | 0 |
| a | 0 | **2** | **1** | **0** | -1 | -1 | -1 |
| c | 0 | 1 | 1 | 0 | **2** | **1** | 0 |
| e | 0 | 0 | 0 | 0 | 0 | 1 | 1 | **3** |

(b) Needleman-Wunsch measure

Table 1: Matrices for computing Smith-Waterman and Needleman-Wunsch scores for sequences **x**=abacde and **x′**=ace, a gap $G = 1$, substitution score $d(x_i, x'_j) = 2$ for $x_i = x'_j$, and $d(x_i, x'_j) = -1$ for $x_i \neq x'_j$.

and $x'_1 \ldots x'_j$. The cell with the largest value in the matrix contains the Smith-Waterman score.

The Needleman-Wunsch measure, which searches for global alignments, is defined similarly, except for the fact that the cells in a matrix can contain negative scores:

$$\underset{\substack{1 \leq i \leq n, \\ 1 \leq j \leq m}}{\text{NW}} (i, j) = \max \begin{cases} \text{NW}(i-1, j-1) + d(x_i, x'_j) \\ \text{NW}(i-1, j) - G \\ \text{NW}(i, j-1) - G \end{cases} \tag{10}$$

The Smith-Waterman measure can be seen as a modification of the Needleman-Wunsch method. By disallowing negative scores in a matrix, the regions of high dissimilarity are avoided and, as a result, local alignments are preferred. Moreover, while the Needleman-Wunsch score equals the largest value in the last column or last row, the Smith-Waterman similarity score corresponds to the largest value in the matrix.

Let us reconsider Example 2 and show how the global and local alignments for alignments for two sequences **x**=abacde and **x′**=ace are obtained. To arrive at actual alignments, one has to set the gap parameter $G$ and the substitution scores. Assume we use the following settings: a gap $G = 1$, substitution score $d(x_i, x'_j) = 2$ for $x_i = x'_j$, and $d(x_i, x'_j) = -1$ for $x_i \neq x'_j$. These values have been chosen for illustrative purpose only, but in a realistic case, e.g., alignment of protein sequences, the choice of the substitution scores is usually motivated by biological evidence. For gapping, Smith and Waterman (1981) suggested to use a gap value which is at least equal to the difference between a match ($d(x_i, x'_j)$, $x_i = x'_j$) and a mismatch ($d(x_i, x'_j)$, $x_i \neq x'_j$). Then, the Smith-Waterman and Needleman-Wunsch similarity scores between **x** and **x′** can be calculated according to Equation 9 and Equation 10 as given in Table 1.

First, the first row and the first column in the matrix are initialized to 0. Then, the matrix is filled in by computing the maximum score for each cell as defined in Equation 9 and Equation 10. The score of the best local alignment is equal to the largest element in the matrix (5), and the Needleman-Wunsch score is 3. Note that it is possible to trace back which steps are taken to arrive at the final alignment (the cells in boldface). A left-right step corresponds to an insertion, a top-down step to a deletion (these lead to gaps), and a diagonal step implies an alignment of two sequences' elements.

Since we prefer to use local alignments on dependency paths, a natural choice would be to use the Smith-Waterman measure as a kernel function. However, Saigo et al. (2004) observed that the Smith-Waterman measure may not result in a valid kernel because it may not be positive semi-definite. They give a definition of the LA kernel, which states that two sequences are similar if they have many local alignments with high scores, as in Equation 11.

$$k_L(\mathbf{x}, \mathbf{x}') = \sum_{\pi \in A(\mathbf{x}, \mathbf{x}')} e^{\beta \cdot s(\mathbf{x}, \mathbf{x}', \pi)} \tag{11}$$

Here, $s(\mathbf{x}, \mathbf{x}', \pi)$ is a local alignment score and $\beta (\geq 0)$ is a scaling parameter.

To define the LA kernel $k_L$ (as in Equation 11) for two sequences $\mathbf{x}$ and $\mathbf{x}'$, it is needed to take into account all transformation operations that are used in local alignments. First, one has to define a kernel on elements that corresponds to individual alignments, $k_a$. Second, since this type of alignment allows gaps, there should be another kernel for gapping, $k_g$. Last but not least, recall that by local alignments only parts of the sequences may be aligned, and some elements of $\mathbf{x}$ and $\mathbf{x}'$ may be left out. These elements do not influence the alignment score and a kernel used in these cases, $k_0$, can be set to a constant, $k_0(\mathbf{x}, \mathbf{x}') = 1$. Finally, the LA kernel is a composition of several kernels ($k_0$, $k_a$, and $k_g$), which is in the spirit of convolution kernels (Haussler, 1999).

According to Saigo et al. (2004), similarity of the aligned sequences' elements ($k_a$ kernel) is defined as follows:

$$k_a(\mathbf{x}, \mathbf{x}') = \begin{cases} 0 & \text{if } |\mathbf{x}| \neq 1 \text{ or } |\mathbf{x}'| \neq 1 \\ e^{\beta \cdot d(\mathbf{x}, \mathbf{x}')} & \text{otherwise} \end{cases} \tag{12}$$

If either $\mathbf{x}$, or $\mathbf{x}'$ has more than one element, this kernel would result in 0. Otherwise, it is calculated using the substitution score $d(\mathbf{x}, \mathbf{x}')$ of $\mathbf{x}$ and $\mathbf{x}'$. This score reflects how similar two sequences' elements are and, depending on the domain, can be computed using prior knowledge from the given domain.

The 'gapping' kernel is defined similarly to the alignment kernel in Equation 12, whereby the scaling parameter $\beta$ is preserved, but the gap penalties are used instead of a similarity function between two elements:

$$k_g(\mathbf{x}, \mathbf{x}') = e^{\beta(g(|\mathbf{x}|) + g(|\mathbf{x}'|))} \tag{13}$$

Here, $g$ stands for the gap function. Naturally, for a gap of length 0 this function returns zero. For gaps of length $n$, it is reasonable to define a gap in terms of a gap opening $o$ and a gap extension $e$, $g(n) = o + e * (n - 1)$. In this case it is possible to decide whether longer gaps should be penalized more than the shorter ones, and how much. For instance, if there are three consecutive gaps in the alignment, the first gap is counted as a gap opening, and the other two as a gap extension. If in consecutive gaps (i.e., gaps of length $n > 1$) each gap is of equal importance, the gap opening has to be equal to the gap extension. If, however, the length of gaps does not matter, one would prefer to penalize the gap opening more, and to give a little weight to the gap extension.

All these kernels can be combined as follows:

$$k_{(r)}(\mathbf{x}, \mathbf{x}') = k_0 * (k_a * k_g)^{r-1} * k_a * k_0 \qquad (14)$$

In Equation 14, $k_{(r)}(\mathbf{x}, \mathbf{x}')$ stands for an alignment of $r$ elements in $\mathbf{x}$ and $\mathbf{x}'$ with possibly $r - 1$ gaps. Similarity of the aligned elements is calculated by $k_a$, and gapping by $k_g$. Since there could be up to $r - 1$ gaps, this corresponds to the following part of the equation: $(k_a * k_g)^{r-1}$. Further, because there is the $r$th aligned element, one more $k_a$ is added. Given the discussion above, $k_0$ is added to the initial and final part. As follows from Equation 14, if there are no elements in $\mathbf{x}$ and $\mathbf{x}'$ aligned, $k_{(r)}$ equals $k_0$, which is 1. If all elements of $\mathbf{x}$ and $\mathbf{x}'$ are aligned with no gaps, the value of $k_{(r)}$ is $(k_a)^r$.

Finally, the LA kernel is equal to the sum taken over all possible local alignments for sequences $\mathbf{x}$ and $\mathbf{x}'$:

$$k_L(\mathbf{x}, \mathbf{x}') = \sum_{i=0}^{\infty} k_{(i)}(\mathbf{x}, \mathbf{x}') \qquad (15)$$

The results in the biological domain suggest that kernels based on the Smith-Waterman distance are more relevant for the comparison of amino acids than string kernels (Saigo et al., 2006). It is not clear whether this holds when applied to natural language processing tasks. In our view, it could depend on the parameters which are used, such as the substitution matrix and the penalty gaps.

### 3.1.1 Computational complexity

The LA kernel, as many other kernels discussed in Section 2, can be efficiently calculated using dynamic programming. For any two sequences $\mathbf{x}$ and $\mathbf{x}'$, of length $n$ and $m$ respectively, its complexity is proportional to $n \times m$. Additional costs may come from the substitution matrix, which, unlike in the biomedical domain, can become very large. However, the look-up of the substitution scores can be done in an efficient manner as well, which leads to fast kernel computation. For instance, calculating a kernel matrix for the largest data set used in this paper, `AImed` (3,763 instances), takes 805 seconds on a 2.93 GHz Intel(R) Core(TM)2 machine.

## 3.2 Designing a Local Alignment Kernel for Relation Extraction

The Smith-Waterman measure is based on transformations, in particular deletions of elements that are different between strings. However, elements that are different may still be similar to some degree. These similarities can be used as part of the similarity measure. For example, if two elements are words that are different but that are synonyms, then we count them as less different than when they are completely unrelated. We will call these similarities "substitution scores" (Equation 12) and define them in two different ways: on the basis of distributional similarity and on the basis of semantic relatedness in an ontology. For Example 1 we would like to be able to infer that 'Brcko' is similar to 'Beijing', even though these two words do not match exactly. Furthermore, if we have phrases "his arrival in Beijing" and "his arrival in January", then we would like our kernel to say that 'Brcko' is more similar to 'Beijing' than to 'January'. The use of such information as prior knowledge makes it possible to measure similarity between two words, one in the test set and the other in the training set, even if they do not match exactly. Below we review two types of measures that are based on statistical distributions and on relatedness in WordNet.

### 3.2.1 Distributional Similarity Measures

There are a number of distributional similarity measures proposed over the years, including *Cosine*, *Dice* and *Jaccard* coefficients. Distributional similarity measures have been extensively studied before (Lee, 1999; Weeds, Weir, & McCarthy, 2004). The main hypothesis behind distributional measures is that words occurring in the same context should have similar meaning (Firth, 1957). Context can be defined either using proximity in text, or employing grammatical relations. In this paper, we use the first option where context is a sequence of words in text and its length is set in advance.

| Measure | Formula |
|---------|---------|
| *Cosine* | $d(x_i, x'_j) = \frac{\sum_c P(c\|x_i) \cdot P(c\|x'_j)}{\sqrt{\sum_c P(c\|x_i)^2 \sum_c P(c\|x'_j)^2}}$ |
| *Dice* | $d(x_i, x'_j) = \frac{2 \cdot F(x_i) \cap F(x'_j)}{F(x_i) \cup F(x'_j)}$ |
| *L2* | $d(x_i, x'_j) = \sqrt{\sum_c (P(c\|x_i) - P(c\|x'_j))^2}$ |

Table 2: A list of functions used to estimate distributional similarity measures.

We have chosen the following measures: *Dice*, *Cosine* and *L2 (Euclidean)* whose definitions are given in Table 2. In the definition of *Cosine* and *L2*, it is possible to use either frequency counts or probability estimates derived from unsmoothed relative frequencies. Here, we adopt the definitions given by Lee (1999), which are based on probability estimates $P$. Recall that $\mathbf{x}$ and $\mathbf{x}'$ are two sequences we would wish to compare, with their corresponding elements $x_i$ and $x'_j$. Further, $c$ stands for a context. In the definition of the *Dice* coefficient, $F(x_i) = \{c : P(c|x_i) > 0\}$. We are mainly interested in symmetric measures ($d(x_i, x'_j) = d(x'_j, x_i)$) because a symmetric positive semi-definite matrix is required by kernel methods. The *Euclidean* measure as defined in Table 2 does not necessarily vary from 0 to 1. For this reason, given a list of pairs of words $(x_i, x'_j)$ where $x_i$ is fixed and $j = 1, \ldots, s$ with their corresponding *L2* score, the maximum value $\max_j d(x_i, x'_j)$ is detected and used to normalize all scores on the list. Furthermore, unlike *Dice* and *Cosine*, which return 1 in the case two words are equal, the *Euclidean* score equals 0. In the next step, we substract the obtained normalized value from 1 to ascertain that all scores are within an interval $[0, 1]$ and the largest value (1) is assigned to identical words. In our view, this procedure will make a comparison of the selected distributional similarity measures with respect to their influence on the LA kernel more transparent.

Distributional similarity measures are very suitable if no other information is available. In the case that data is annotated by means of some taxonomy (e.g., WordNet), it is possible to consider measures defined over this taxonomy. Availability of hand-crafted resources, such as WordNet, that comprise various relations between concepts, enables making distinctions between different concepts in a subtle way.

### 3.2.2 WordNet Relatedness Measures

For generic relations, the most commonly used resource is WordNet (Fellbaum, 1998), which is a lexical database for English. In WordNet, words are grouped together in synsets where a synset "consists of a list of synonymous words or collocations (e.g., 'fountain pen'), and pointers that describe the relations between this synset and other synsets" (Fellbaum, 1998). WordNet can be employed for different purposes such as studying semantic constraints for certain relation types (Girju, Badulescu, & Moldovan, 2006; Katrenko & Adriaans, 2008), or enriching the training set (Giuliano et al., 2007; Nulty, 2007).

To compare two concepts given their synsets $c_1$ and $c_2$ we use five different measures that have been proposed in the past years. Most of them rely on the notions of the length of the shortest path between two concepts $c_1$ and $c_2$, $len(c_1, c_2)$, the depth of a node in the WordNet hierarchy (which is equal to the length of the path from the root to the given synset $c_i$), $dep(c_i)$, and a least common subsumer (or lowest super-ordinate) between $c_1$ and $c_2$, $lcs(c_1, c_2)$, which in turn is a synset. To the measures that are exclusively based on these notions belong conceptual similarity proposed by Palmer and Wu (1995) ($sim_{wup}$ in Equation 16) and the formula of scaled semantic similarity introduced by Leacock and Chodorow (1998) ($sim_{lch}$ in Equation 17). [1] The major difference between them lies in the fact that $sim_{lch}$ does not consider the least common subsumer of $c_1$ and $c_2$ but uses the maximum depth of the WordNet hierarchy instead. Conceptual similarity ignores this and focuses on the subhierarchy that includes both synsets.

$$sim_{wup}(c_1, c_2) = \frac{2 * dep(lcs(c_1, c_2))}{len(c_1, lcs(c_1, c_2)) + len(c_2, lcs(c_1, c_2)) + 2 * dep(lcs(c_1, c_2))} \qquad (16)$$

$$sim_{lch}(c_1, c_2) = -\log \frac{len(c_1, c_2)}{2 * \max_{c \in WordNet} dep(c)} \qquad (17)$$

Aiming at combining information from several sources, Resnik (1995) introduced yet another measure that is grounded in information content ($sim_{res}$ in Equation 18). Intuitively, if two synsets $c_1$ and $c_2$ are located deeper in the hierarchy and the path from one synset to another is short, they should be similar. If the path between two synsets is long and their least common subsumer is placed relatively close to the root, this indicates that the synsets

---

1. In all equations of similarity measures defined over WordNet, subscripts refer to the similarity measure itself (e.g., $lch$, $wup$ in $sim_{lch}$ and in $sim_{wup}$, respectively) $c_1$ and $c_2$ do not have much in common. To quantify this intuition, it is necessary to derive a probability estimate for $lcs(c_1, c_2)$ which can be done by employing existing corpora. More precisely, $p(lcs(c_1, c_2))$ stands for the probability of encountering an instance of a concept $lcs(c_1, c_2)$.

$$sim_{res}(c_1, c_2) = -\log p(lcs(c_1, c_2)) \tag{18}$$

One of the biggest shortcomings of Resnik's method is the fact that only the least common subsumer appears in Equation 18. One can easily imagine a full-blown hierarchy where the relatedness of the concepts subsumed by the same $lcs(c_i, c_j)$ can heavily vary. In other words, by using $lcs$ only, one is not able to make subtle distinctions between two pairs of concepts that share the least common subsumer. To overcome this, Jiang and Conrath (1997) proposed a solution that takes into account information about the synsets being compared ($sim_{jcn}$ in Equation 19). By comparing Equation 19 against Equation 18, we will notice that now the equation incorporates not only the probability of encountering $lcs(c_1, c_2)$, but also the probability estimates for $c_1$ and $c_2$.

$$sim_{jcn}(c_1, c_2) = 2 \log p(lcs(c_1, c_2)) - (\log p(c_1) + \log p(c_2)) \tag{19}$$

Lin (1998) defined the similarity between two concepts using how much commonality and differences between them are involved. Similarly to the two previous approaches, he uses information theoretic notions and derives the similarity measure $sim_{lin}$ given in Equation 20.

$$sim_{lin}(c_1, c_2) = \frac{2 * \log p(lcs(c_1, c_2))}{\log p(c_1) + \log p(c_2)} \tag{20}$$

In the past, semantic relatedness measures were evaluated on different NLP tasks (Budanitsky & Hirst, 2006; Ponzetto & Strube, 2007) and it can be concluded that no measure performs the best for all problems. In our evaluation, we use semantic relatedness for the validation of generic relations and study in depth how they contribute to the final results.

### 3.2.3 Substitution Matrix for Relation Extraction

Until now, we have discussed two possible ways of calculating the substitution score $d(\cdot, \cdot)$, by using either distributional similarity measures, or measures defined on WordNet. However, dependency paths which are generated by parsers may contain not only words (or lemmata), but also syntactic functions such as subjects, objects, modifiers, and others. To take this into account, we revise the definition of $d(\cdot, \cdot)$. We assume sequences $\mathbf{x} = x_1 x_2 \ldots x_n$ and $\mathbf{x}' = x_1' x_2' \ldots x_m'$ to contain words ($x_i \in W$ where $W$ refers to a set of words) and syntactic functions accompanied by direction ($x_i \notin W$). The elements of $W$ are unique words (or lemmata) which are found in the dependency paths, for instance, for the paths 'his $\rightarrow$ actions $\leftarrow$ in $\leftarrow$ Brcko' and 'his $\rightarrow$ arrival $\leftarrow$ in $\leftarrow$ Beijing' in Example (1) in Section 2.3.5, $W=$ {his, actions, in, Brcko, arrival, Beijing}. The dependency paths we use in the present work include information on syntactic functions, for instance 'awareness $\overset{prep\_from}{\leftarrow}$ come $\overset{nsubj}{\rightarrow}$ joy'. In this case, $W=$ {awareness, come, joy} and $\bar{W} = \{\overset{prep\_from}{\leftarrow}, \overset{nsubj}{\rightarrow}\}$.

Then,

$$d'(x_i, x'_j) = \begin{cases} d(x_i, x'_j) & x_i, x'_j \in W \\ 1 & x_i, x'_j \notin W \ \& \ x_i = x'_j \\ 0 & x_i, x'_j \notin W \ \& \ x_i \neq x'_j \\ 0 & x_i \in W \ \& \ x'_j \notin W \\ 0 & x_i \notin W \ \& \ x'_j \in W \end{cases} \tag{21}$$

Equation (21) states that whenever the element $x_i$ of the sequence $\mathbf{x}$ is compared against the element $x'_j$ of the sequence $\mathbf{x}'$, their substitution score is equal either to (i) the similarity score in the case both elements are words (lemmata), or to (ii) 1, if both elements are the same syntactic function, or to (iii) 0, in any other case.

As follows from our discussion on similarity measures above, there are two ways to define $d(x_i, x'_j)$, using either distributional similarity between $x_i$ and $x'_j$ (Section 3.2.1), or their WordNet similarity, provided that they are annotated with WordNet synsets (Section 3.2.2).

## 4. Experimental Set-up

In this section, we describe the data sets that we have used in the experiments and provide information on the data collections used for estimating distributional similarity.

### 4.1 Data

To evaluate the performance of the LA kernel, we consider two types of text data, domain-specific data, which comes from the biomedical domain and generic or domain-independent data which represents a variety of well-known and widely used relations such as Part-Whole and Cause-Effect.

Like other work, we extract a dependency path between two nodes corresponding to the arguments of a binary relation. We also assume that each analysis results in a tree and since it is an acyclic graph, there exists a unique path between each pair of nodes. We do not consider, however, other structures that might be derived from the full syntactic analysis as in, for example, subtrees (Moschitti, 2006).

#### 4.1.1 Biomedical Relations

**Corpora** We use three corpora that come from the biomedical field and contain annotations of either interacting proteins - BC-PPI[2] (1,000 sentences), AImed (Bunescu & Mooney, 2005b) or the interactions among proteins and genes LLL (77 sentences in the training set and 87 in the test set, Nédellec, 2005). The BC-PPI corpus was created by sampling sentences from the BioCreAtIve challenge, the AImed corpus was sampled from the Medline collection. The LLL corpus was composed by querying Medline with the term *Bacillus subtilis*. The difference among all three corpora lies in the directionality of interactions. As Table 3 shows, relations in the AImed corpus are strictly symmetric, in LLL they are asymmetric and BC-PPI contains both types. The differences in the number of training instances for the AImed corpus can be explained by the fact that they correspond to the dependency paths between named entities. If parsing fails or produces several disconnected graphs per sentence, no dependency path is extracted.

---

2. Available from `http://www2.informatik.hu-berlin.de/~hakenber/`.

| Parser | Data set | #examples | #pos | direction |
|--------|----------|-----------|------|-----------|
| LinkParser | `LLL (train)` | 618 | 153 | asymmetric |
| LinkParser | `LLL (test)` | 476 | 83 [a] | asymmetric |
| Stanford | `BC-PPI` | 664 | 250 | mixed |
| Stanford | `AImed` | 3763 | 922 | symmetric |
| Enju | `AImed` | 5272 | 918 | symmetric |

a. Even though the actual annotations for the test data are not given, the number of interactions in the test data set is provided by the `LLL` organizers.

Table 3: Statistics of the biomedical data sets `LLL`, `BC-PPI`, and `AImedd`. In this table, #pos stands for the number of positive examples per data set and #examples indicates the number of examples in total.

The goal of relation extraction in all three cases is to output all correct interactions between biomedical entities (genes and proteins) that can be found in the input data. The biomedical entities are already provided, so there is no need for named entity recognition.

There is a discrepancy between the training and the test sets used for the LLL challenge. Unlike the training set, where each sentence has an example of at least one interaction, the test set contains sentences with no interaction. The organizers of the `LLL` challenge distinguish between sentences with and without coreferences. Sentences with coreferences are usually appositions, as shown in one of the examples below. The first sentence in (4.1.1) is an example of a sentence without coreferences (with interaction between 'ykuD' and 'SigK'), whereas the second one is a sentence with coreference (with interaction between 'spoIVA' and 'sigmaE'). More precisely, 'spoIVA' refers to the phrase 'one or more genes' which are known to interact with 'sigmaE'. We can therefore infer that 'spoIVA' interacts with 'sigmaE'. Sentences without coreferences form a subset, which we refer to as `LLL-nocoref`, and sentences with coreferences are part of the separate subset `LLL-coref`.

(22)  *ykuD* was transcribed by *SigK* RNA polymerase from T4 of sporulation.

(23)  Finally, we show that proper localization of SpoIVA required the expression of one or more genes which, like *spoIVA*, are under the control of the mother cell transcription factor *sigmaE*.

It is assumed here that relations in the sentences with coreferences are harder to recognize. To show how the LA kernel performs on both subsets, we report the experimental results on the full set of test data (`LLL-all`), and on its subsets (`LLL-coref` and `LLL-nocoref`).

**Syntactic analysis**  We analyzed the `BC-PPI` corpus with the Stanford parser. The `LLL` corpus has already been preprocessed by the LinkParser and its output was checked by experts. To enable comparison with the previous work, we used the `AImed` corpus parsed by the Stanford parser [3] and by the Enju parser [4] (which exactly correspond to the input in the experiments by Erkan et al., 2007 and Sætre et al., 2008). Unlike the Stanford parser, Enju is based on a Head-driven Phrase Structure Grammar (HPSG). The output of the Enju parser can be presented in two ways, either as predicate argument structure or as a phrase structure tree. Predicate argument structures describe relations between words in a sentence, while phrase structure presents a sentence structure in the form of clauses and phrases. In addition, Enju was trained on the GENIA corpus and includes a model for parsing biomedical texts.

(24) Cbf3 contains three proteins, Cbf3a, Cbf3b and Cbf3c.

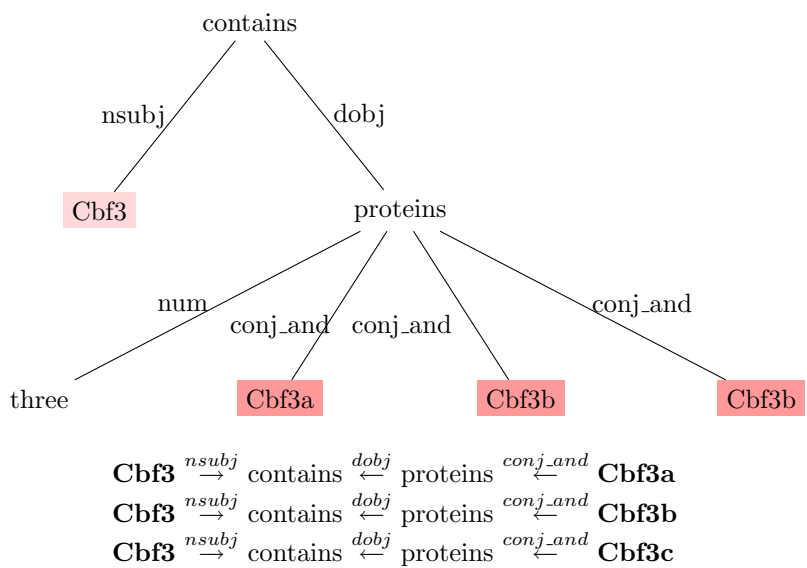

Figure 1: Stanford parser output and representation for Example (24).

Figure 1 shows a dependency tree obtained by the Stanford parser for the sentence in (24). This sentence mentions three interactions among proteins, more precisely, between 'Cbf3' and 'Cbf3a', 'Cbf3' and 'Cbf3b', and 'Cbf3' and 'Cbf3c'. All three dependency paths contain words (lemmata) and syntactic functions (such as *subj* for a subject) plus the direction of traversing the tree. Figure 2 presents the output for the same sentence provided by the Enju parser. The upper part refers to the phrase structure tree and the lower part shows the paths extracted from the predicate argument structure. The two parsers clearly differ in their output. First, the Stanford parser conveniently generates the same paths for all three interaction pairs while the Enju analyzer does not. Second, the output of the Stanford parser excludes prepositions or conjunctions that are attached to the syntactic functions whereas the Enju analyzer lists them in the parsing results. Such differences

---

lead to different input sequences that are later fed into the LA kernel. Consequently, the variations in input may translate into differences in the final performance.

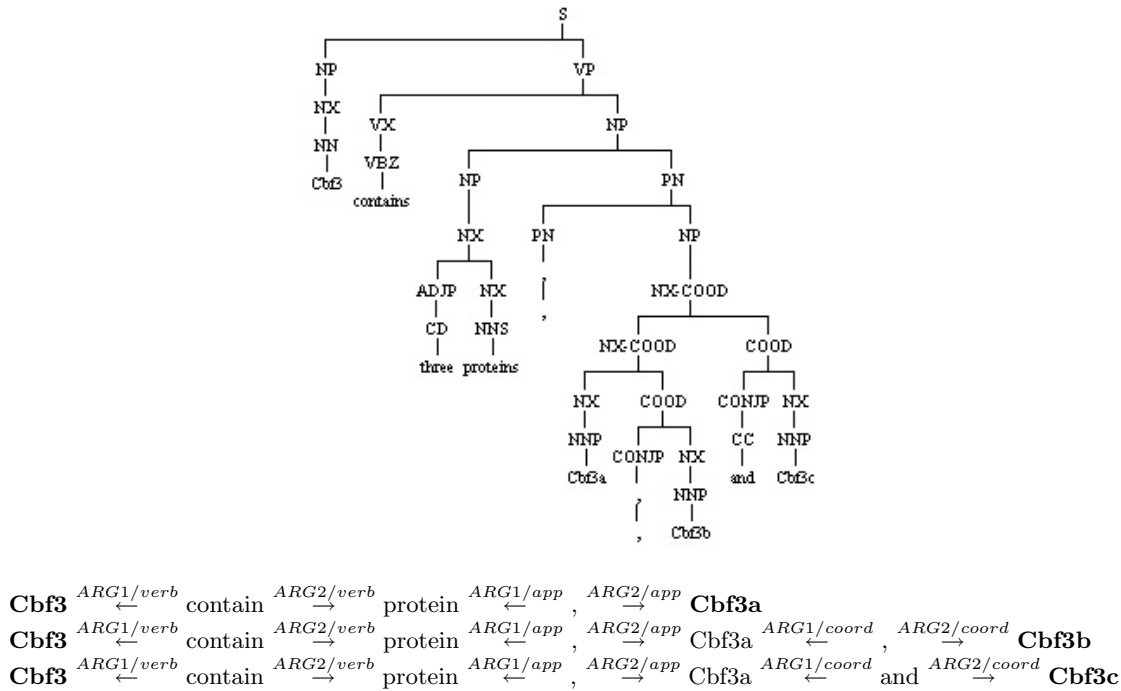

$$\mathbf{Cbf3} \overset{ARG1/verb}{\leftarrow} \text{contain} \overset{ARG2/verb}{\rightarrow} \text{protein} \overset{ARG1/app}{\rightarrow} \overset{ARG2/app}{\rightarrow} \mathbf{Cbf3a}$$

$$\mathbf{Cbf3} \overset{ARG1/verb}{\leftarrow} \text{contain} \overset{ARG2/verb}{\rightarrow} \text{protein} \overset{ARG1/app}{\rightarrow} , \overset{ARG2/app}{\rightarrow} \text{Cbf3a} \overset{ARG1/coord}{\leftarrow} \overset{ARG2/coord}{\rightarrow} \mathbf{Cbf3b}$$

$$\mathbf{Cbf3} \overset{ARG1/verb}{\leftarrow} \text{contain} \overset{ARG2/verb}{\rightarrow} \text{protein} \overset{ARG1/app}{\leftarrow} , \overset{ARG2/app}{\rightarrow} \text{Cbf3a} \overset{ARG1/coord}{\rightarrow} \text{and} \overset{ARG2/coord}{\rightarrow} \mathbf{Cbf3c}$$

Figure 2: Enju's output and representation for Example (24).

In addition, in most work employing `AImed`, the dependency paths such as these in Figure 1 and Figure 2 are preprocessed in the following way. The actual named entities that are the arguments of the relation are replaced by a label, e.g. PROTEIN. Consequently, the first path in Figure 1 becomes 'PROTEIN $\overset{nsubj}{\rightarrow}$ contains $\overset{dobj}{\leftarrow}$ proteins $\overset{conj\_and}{\leftarrow}$ PROTEIN'. To be able to compare our results on `AImed` with the performance reported in the work of Erkan et al. (2007) and Sætre et al. (2008), we use exactly the same dependency paths with argument labels. However, to study whether using labels instead of actual named entities has an impact on the final results for the `LLL` data set, we carry out two experiments. In the first one, the dependency paths contain named entities, whereas in the second they contain labels. The second experiment is referred to by adding a word 'LABEL' to its name (as `LLL-all-LABEL` in Table 7).

### 4.1.2 Generic Relations

The second type of relations that we consider are generic relations. Their arguments are sometimes annotated using external resources such as WordNet, which makes it possible to use semantic relatedness measures defined over them. An example of such an approach is data used for the `SemEval-2007` challenge, "Task 04: Classification of Semantic Relations between Nominals" (Girju et al., 2009).

The goal of Task 4 was to classify seven semantic relations (Cause - Effect, Instrument - Agency, Product - Producer, Origin - Entity, Theme - Tool, Part - Whole and Content - Container), whose examples were collected from the Web using some predefined queries. In other words, given a set of examples and a relation, the expected output would be a binary classification of whether an example belongs to the given relation or not. The arguments of the relation were annotated by synsets from the WordNet hierarchy, as in Figure 3. Given this sentence and a pair (spiritual awareness, joy) with the corresponding synsets joy%1:12:00 and awareness%1:09:00, this would mean that a classifier has to decide whether this pair is an example of the Cause-Effect relation. This particular sentence was retrieved by quering the Web with the phrase "joy comes from *". The synsets were manually selected from the WordNet hierarchy. There are seven semantic relations used in this challenge, which gives seven binary classification problems.

---

Genuine <e1>joy</e1> comes from <e2>spiritual awareness</e2> on life and an absolute clarity of direction, living for a purpose.

WordNet(e1) = "joy%1:12:00", WordNet(e2) = "awareness%1:09:00",
Query: "joy comes from *", Cause-Effect(e2, e1) = true

---

Figure 3: An annotated example of Cause - Effect from the `SemEval-2007`, Task 4 training data set.

| relation type | #examples (train) | #pos (train) | #examples (test) | direction |
|---|---|---|---|---|
| Origin - Entity | 140 | 54 | 81 | asymmetric |
| Product - Producer | 140 | 85 | 93 | asymmetric |
| Theme - Tool | 140 | 58 | 71 | asymmetric |
| Instrument - Agency | 140 | 71 | 78 | asymmetric |
| Part - Whole | 140 | 65 | 72 | asymmetric |
| Content - Container | 140 | 65 | 74 | asymmetric |
| Cause - Effect | 140 | 73 | 80 | asymmetric |

Table 4: Distribution of the `SemEval-2007`, Task 4 examples (training and test data), where #pos stands for the number of positive examples per data set and #examples indicates the number of examples in total.

**Syntactic analysis**   To generate dependency paths, all seven data sets used in `SemEval - 2007`, Task 4, were analyzed by the Stanford parser. The dependency path for the sentence in Figure 3 is given in (25).

(25)  awareness#n#1 $\overset{prep\_from}{\longleftarrow}$ come $\overset{nsubj}{\longrightarrow}$ joy#n#1

Here, words annotated with WordNet have their PoS tag attached, followed by the sense. For instance, 'awareness' is a noun and in the current context its first sense is used, which corresponds to 'awareness#n#1'.

## 4.2 Substitution Matrix

To build a substitution matrix for the LA kernel, we use either distributional similarity or WordNet semantic relatedness measures. For a data set of dependency paths, which contains $t$ unique elements (words and syntactic functions), the size of the matrix is $t \times t$. If $k$ elements out of $t$ are words, the number of substitution scores to be computed by distributional similarity (or semantic relatedness) measures equals $k(k + 1)/2$. This is due to the fact that the measures we use are symmetric. The substitution matrix is built for each corpus we used in the experiments, which results in three substitution matrices for the biomedical domain (for `BC-PPI`, `LLL`, and `AImed`) and seven substitution matrices for generic relations. In what follows, we discuss the settings which were used for calculating the substitution matrix in more detail.

Distributional similarity can be estimated either by using contextual information (Ó Séaghdha & Copestake, 2008), or by exploring grammatical relations between words (Lee, 1999). In this work we opt for contextual information. This is motivated by the presence of words belonging to different parts of speech in the dependency paths. For instance, even though, according to dependency grammar theory (Mel'čuk, 1988), adjectives do not govern other words, they may still occur in the dependency paths. In other words, even if parsing does not fail, it may produce unreliable syntactic structures. To be able to compare words of any part of speech, we have decided to estimate distributional similarity based on contextual information, rather than on grammatical relations.

While computing distributional similarity, it may happen that a given word $x_i$ does not occur in the corpus. To handle such cases, we always set $d(x_i, x_i) = 1$ (the largest possible similarity score), and $d(x_i, x'_j) = 0$ when $x_i \neq x'_j$ (the lowest possible similarity score).

### 4.2.1 Biomedical domain

To estimate distributional similarity for the biomedical domain, we use the `TREC 2006 Genomics` collection (Hersch, Cohen, Roberts, & Rakapalli, 2006) which contains 162,259 documents from 49 journals. All documents have been preprocessed by removing HTML-tags, citations in the text and reference sections and stemmed by the Porter stemmer (van Rijsbergen, Robertson, & Porter, 1980). Furthermore, the query-likelihood approach with Dirichlet smoothing (Chen & Goodman, 1996) is used to retrieve document passages given a query. All passages are ranked according to their likelihood of generating the query. Dirichlet smoothing is used to avoid zero probabilities and poor probability estimates (which may happen when words do not occur in the documents). All $k$ unique words occurring in the set of dependency paths sequences are fed as queries to collect a corpus for estimating similarity. Immediate context surrounding each pair of words is used to calculate the distributional similarity of these words. We set the context window to $\pm 2$ (2 tokens to the right and 2 tokens to the left of a word in focus) and do not perform any kind of further preprocessing such as PoS tagging.

### 4.2.2 Generic relations

For generic relations, we use all WordNet relatedness measures described in Section 3.2.2. We have already shown that the WordNet relatedness measures work only on synsets, which assumes that all words have to be manually annotated with information from WordNet. Since this is done only for the relations' arguments (see the example in Figure 3), and for no other words in sentences (and, correspondingly, in the dependency paths), we build a substitution matrix as follows. For any two words annotated with WordNet, their substitution score equals a value returned by a relatedness measure being used. For any other word pair, it equals 1 whenever the words are identical, and 0 otherwise. [5] For example, if we consider the words in the dependency path in (25) and the Wu-Palmer (*wup*) relatedness measure, the substitution scores that we obtain are as follows:

| | |
|---|---|
| d(awareness#n#1, awareness#n#1) = 1 | d(awareness#n#1, prep_from↑) = 0 |
| d(awareness#n#1, come) = 0 | d(awareness#n#1, nsubj↓) = 0 |
| d(awareness#n#1, joy#n#1) = 0.35 | d(prep_from↑, prep_from↑) = 1 |
| d(prep_from↑, come) = 0 | d(prep_from↑, nsubj↓) = 0 |
| d(prep_from↑, joy#n#1) = 0 | d(come, come) = 1 |
| d(come, nsubj↓) = 0 | d(come, joy#n#1) = 0 |
| d(nsubj↓, nsubj↓) = 1 | d(nsubj↓, joy#n#1) = 0 |
| d(joy#n#1, joy#n#1) = 1 | |

Figure 4: The substitution scores for the dependency path in (25) using *wup* measure. Syntactic relations (prep_from, subj) are accompanied by the direction of the dependency tree traversal (either ↑ or ↓).

In the dependency path (25), there are 5 unique elements ($t$), 2 of which are annotated with WordNet synsets ($k$). Consequently, there are $5*6/2 = 15$ substitution scores in total, 3 of which are computed using WordNet relatedness.

To compute WordNet relatedness, we use the `WordNet::Similarity` package for WordNet 3.0 (Pedersen, Patwardhan, & Michelizzi, 2004).

## 4.3 Baselines and Kernel Settings

In this section, we discuss two baselines and kernel settings.

### 4.3.1 Baselines

To test how well local alignment kernels perform compared to kernels proposed in the past, we implemented the shortest path kernel described in the work of Bunescu and Mooney (2005a) (Section 2.3.5) as one of the baselines (Baseline I). This method seems to be the most natural choice because it operates on the same data structures (dependency paths). Similarly to Bunescu and Mooney's (2005a) work, in our experiments we use lemma, part of speech tag and direction, but we do not consider entity type or negative polarity of items.

---

5. This also applies to the cases when the relation arguments could not have been annotated with WordNet information.

The choice of the LA kernel in this paper was motivated not only by its ability to compare sequences in a flexible way, but also because of the possibility to explore additional information (not present in the training set) via a substitution matrix. The other baseline, Baseline II, is used to test whether the choice of similarity measures affects the results. In this case, the substitution scores $d(\cdot, \cdot)$ are not calculated using distributional similarity or WordNet relatedness, but generated randomly within the interval $[0, 1]$.

### 4.3.2 Kernel settings

The kernels we compute are used together with the support vector machine tool LibSVM (Chang & Lin, 2001) to detect hyperplanes separating positive examples from negative ones. Before plugging all kernel matrices for 10-fold cross-validation into LibSVM, they are normalized as in Equation 26.

$$k(x^{'}, y^{'}) = \frac{k(x, y)}{\sqrt{k(x, x)k(y, y)}} \tag{26}$$

To handle imbalanced data sets (most notably `AImed` and `BC-PPI`), the examples are weighted using inverse-class probability (i.e. all training examples of class $A$ are weighted $1/\text{prob}(A)$ where $\text{prob}(A)$ is the fraction of training examples with class $A$). All significance tests were done using a two-tailed paired $t$-test with confidence level 95% ($\alpha = 0.05$).

In addition, in all experiments we tuned the penalty parameter $C$ (Equation 4) in the range $(2^{-6}, 2^{-4}, \ldots, 2^{12})$.

To use the LA kernel, one has to set the following parameters: the gap opening cost, the gap extension cost, and the scaling parameter $\beta$. In our cross-validation experiments, the gap opening cost is set to 1.2, the extension cost to 0.2 and the scaling parameter $\beta$ to 1. The choice of the scaling value was motivated by the experiments on amino acids in the biological domain (Saigo et al., 2004). After initial experiments, we present here a further study where the parameter values are varied.

## 5. Experiment I: Domain-Specific Relations

The goal of this evaluation is to study the behavior of the LA kernel on domain-specific relations in the biomedical domain. In this section, we report on the experiments conducted on three biomedical corpora using the LA kernel based on the distributional similarity measures, two baselines and results published previously (e.g., using the graph kernel by Airola et al., 2008 or the tree kernel by Sætre et al., 2008). To the best of our knowledge, string kernels have not been applied to dependency paths yet. However, a gap-weighted string kernel (described in Section 2) also allows gapping and can be thus compared to the LA kernel. To test how Lodhi et al.'s (2002) kernel performs on dependency paths, we use it on all three corpora. We have not tuned parameters of this string kernel and set the length of subsequences to 4 and the decay factor $\lambda$ to 0.5. [6]

## 5.1 `LLL` and `BC-PPI` Data Sets

This subsection presents results on two biomedical data sets, `BC-PPI` and `LLL`. Whenever possible, we also discuss the performance previously reported in the literature.

The 10-fold cross-validation results on the `BC-PPI` corpus are presented in Table 5 and on the `LLL` training data set in Table 6. The LA kernel based on the distributional similarity measures (LA-Dice, LA-Cosine and LA-L2) performs significantly better than the two baselines. Recall that Baseline I corresponds to the shortest path approach (Section 2.3.5) and Baseline II is the LA kernel with the randomly generated substitution scores. In contrast to Baseline I, it is able to handle sequences of different lengths including gaps. According to Equation 5, a comparison of any two sequences of different lengths results in the 0-score. Nevertheless, it still yields high recall, while precision is much lower. This can be explained by the fact that the shortest path uses PoS tags. Even though two sequences of the same length can be very different, their comparison may still result in a non-zero score, provided that their part of speech tags match. Furthermore, Baseline II suggests that accurate estimation of substitution scores is important for achieving good performance. Baseline II may yield better results than Baseline I, but randomly generated substitution scores degrade the performance.

| Method | Precision | Recall | F-score |
|---|---|---|---|
| LA-Dice | 75.56 | 79.72 | 77.56 |
| LA-Cosine | 76.40 | 80.66 | 78.13 |
| LA-L2 | 77.56 | 79.31 | 78.42 |
| Baseline I | 32.04 | 75.63 | 45.00 |
| Baseline II | 66.36 | 54.48 | 59.80 |
| Gap-weighted string kernel (Lodhi et al., 2002) | 72.00 | 75.31 | 73.62 |

Table 5: 10-fold cross-validation on the `BC-PPI` data set.

At first glance, the choice of the distributional similarity measures does not affect the overall performance yielded by the LA kernel. On the `BC-PPI` data, the method based on the *L2* measure outperforms the methods based on *Dice* (p$\leq$.07) and on *Cosine*, but the differences in the latter case are not significant. No statistically significant differences were observed between the method based on *Dice* and *Cosine*.

In contrast to the `BC-PPI` data set, the kernels which use *Dice* and *Cosine* measures on the `LLL` data set significantly outperform the one based on *L2* (at p$\leq$1.22$\times10^{-7}$ and p$\leq$1.33$\times10^{-6}$, respectively).

On both data sets, the LA method using distributional similarity measures significantly outperforms the baselines. Interestingly, the gap-weighted string kernel by Lodhi et al. (2002) yields good performance too and seems to be a better choice than the subsequence kernel based on shallow linguistic information (Giuliano et al., 2006). Recent work on `LLL` (Fundel, Kueffner, & Zimmer, 2007) employs dependency information but, in contrast to our method, it serves as the representation on which extraction rules are defined. Airola et al. (2008) apply a graph kernel-based approach to extract interactions and use, among others, the `LLL` and `AImed` data sets. As can be seen in Table 6, their method yields results which are comparable to the gap-weighted string kernel on the dependency paths. To the best of our knowledge, the performance achieved by the LA kernel on the `LLL` training set is the highest (in terms of the F-score) among the results which have been reported in the literature.

---

[6]. Lodhi et al. (2002) have mentioned in their paper that "the $F_1$ numbers (with respect to SSK) seem to peak at a subsequence length between 4 and 7".

| Method | Precision | Recall | F-score |
|---|---|---|---|
| LA-Dice | 88.76 | 81.62 | 85.03 |
| LA-Cosine | 88.63 | 82.09 | 85.23 |
| LA-L2 | 86.80 | 75.04 | 80.49 |
| Baseline I | 39.02 | 100.00 | 56.13 |
| Baseline II | 65.82 | 41.32 | 50.76 |
| Graph kernel (Airola et al., 2008) | 72.5 | 87.2 | 76.8 |
| Gap-weighted string kernel (Lodhi et al., 2002) | 83.66 | 71.11 | 76.88 |
| Shallow linguistic kernel (Giuliano et al., 2006) | 62.10 | 61.30 | 61.70 |
| Rule-based method (Fundel et al., 2007) | 68 | 83 | 75 |

Table 6: 10-fold cross-validation on the `LLL-all` training data set.

We also apply our method to the `LLL` test data (Table 7). [7] Even though the performance on the test set is poorer, LA-Dice outperforms both baselines. In addition, the gap-weighted string kernel (Lodhi et al., 2002) seems to perform much worse on the test set. For the LA kernel, precision is high, while recall decreases (and most drastically for the data subset which includes co-references). This might be due to the fact that for some sentences only incomplete parses are generated and, consequently, no dependency paths between the entities are found. For 91 out of 567 possible interaction pairs generated on the test data, there is no dependency path extracted. In contrast, the approach reported by Giuliano et al. (2006) does not make use of syntactic information, and on the data subset without coreferences achieves higher recall.

On the other hand, lower recall can also be caused by using actual names of proteins and genes as arguments. In the work reported before, the relation arguments and other named entities are often replaced by their types (e.g., PROTEIN) and these are used as input for the learning algorithm. We conducted additional experiments using named entity types in the dependency paths, which led to a great improvement in terms of recall and F-score (Table 7, `LLL-coref-LABEL`, `LLL-nocoref-LABEL`, `LLL-coref-LABEL`). Our method clearly outperforms the shallow linguistic kernel and also achieves better results than the best-performing system in the LLL competition ($S_{best}$), which, according to Nédellec (2005), applied Markov logic to the syntactic paths.

---

7. Airola et al. (2008) do not report on the performance on the `LLL` data set and, for this reason, information on the graph all-paths kernel is not included in Table 7.

| Data set | Method | Precision | Recall | F-score |
|---|---|---|---|---|
| `LLL-coref` | LA-Dice | 52.3 | 37.9 | 44.0 |
| `LLL-nocoref` | LA-Dice | 70.7 | 53.7 | 61.0 |
| `LLL-all` | LA-Dice | 72.7 | 48.1 | 57.9 |
| `LLL-all` | Baseline I | 48.6 | 43.3 | 45.8 |
| `LLL-all` | Baseline II | 12.9 | 45.7 | 20.1 |
| `LLL-coref-LABEL` | LA-Dice | 60.0 | 51.7 | 55.5 |
| `LLL-nocoref-LABEL` | LA-Dice | 69.0 | 53.7 | 60.4 |
| `LLL-all-LABEL` | LA-Dice | 74.5 | 53.0 | 61.9 |
| `LLL-coref` | Shallow linguistic kernel (Giuliano et al., 2006) | 29.0 | 31.0 | 30.0 |
| `LLL-nocoref` | Shallow linguistic kernel (Giuliano et al., 2006) | 54.8 | 62.9 | 58.6 |
| `LLL-all` | Shallow linguistic kernel (Giuliano et al., 2006) | 56.0 | 61.4 | 58.6 |
| `LLL-all` | Gap-weighted string kernel (Lodhi et al., 2002) | 56.0 | 16.8 | 25.9 |
| `LLL-all` | $S_{best}$ (Nédellec, 2005) | 60.9 | 46.2 | 52.6 |

Table 7: Results on the `LLL` test data set.

## 5.2 `AImed` Data Set

Yet another data set that we consider is `AImed`. This data set has often been used for experiments on relation extraction in the biomedical domain, which enables comparison with other methods. It should be noted, however, that in this particular case, a corpus is a collection of documents (abstracts). This may lead to two ways of performing 10-fold cross-validation. One possibility lies in randomly splitting data in 10 parts, while the other is to do cross-validation on the level of documents. The experiments we report here are done using the first setting and can be directly compared against the methods described in the work of Sætre et al. (2008), Erkan et al. (2007) and Giuliano et al. (2006). In addition, we use the same dependency paths for the LA kernel as the ones employed by Sætre et al. and Erkan et al.. The results by Airola et al. (2008) and by Bunescu (2007) are obtained by cross-validating on the level of documents.

We conducted experiments by setting the distributional measure to *Dice*, referred to as LA-Dice in Table 8. In the upper part of the table we used dependency paths generated by the Stanford parser and in the lower part those obtained by Enju. As we discussed in Section 2, Erkan et al. (2007) use similarity measures to compare dependency paths, but they do not consider any additional sources whose information can be incorporated into the learning procedure. They, however, experiment with supervised (SVM) and semi-supervised learning (TSVM), where the number of training instances is varied. Table 8 shows the best performance that was achieved by Erkan et al.'s (2007) method. Among models based on SVM, the one with *Cosine* distance, SVM-Cos, yields the best results. In the TSVM setting, the one with the *Edit* measure performs the best. We observe that LA-Dice slightly outperforms both and has, in particular, high precision.

In their work, Sætre et al. (2008) explore several parsers and combinations of features. The features include not only paths from Enju, but also word dependencies generated by data-driven KSDEP parser, and word features. KSDEP parser is based on a probabilistic shift-reduce algorithm (Sagae & Tsujii, 2007). In general, the method by Sætre et al. also uses SVM, but in this case it focuses on tree kernels (discussed in Section 2.3.3). To make a fair comparison, we conducted experiments on the paths obtained by deep syntactic analysis (Enju parser) and compared our scores against Sætre et al.'s (2008) results. In contrast to the previous experiments, we achieve higher recall but lower precision. Overall, the LA kernel yields better performance than the one reported by Sætre et al. However, when different sets of features are combined (parses from Enju and KSDEP plus word features - 'Enju+KSDEP+W' in Table 8), the overall performance can be improved.

| Method | Parser | Precision | Recall | F-score |
|---|---|---|---|---|
| LA-Dice | Stanford | 69.09 | 54.63 | 61.02 |
| Baseline I (Bunescu, 2007) | Collins | 69.08 | 35.00 | 46.46 |
| Baseline II | Stanford | 48.89 | 25.06 | 33.07 |
| SVM-Cos (Erkan et al., 2007) | Stanford | 61.99 | 54.99 | 58.09 |
| TSVM-Edit (Erkan et al., 2007) | Stanford | 59.59 | 60.68 | 59.96 |
| Gap-weighted string kernel (Lodhi et al., 2002) | Stanford | 67.25 | 54.67 | 60.31 |
| LA-Dice | Enju | 71.16 | 46.71 | 56.40 |
| Tree kernel (Sætre et al., 2008) | Enju | 76.0 | 39.7 | 52.0 |
| Tree kernel (Sætre et al., 2008) | Enju+KSDEP+W | 78.1 | 62.7 | 69.5 |
| Graph kernel (Airola et al., 2008) | Charniak-Lease | 52.9 | 61.8 | 56.4 |
| Shallow linguistic kernel (Giuliano et al., 2006) | none | 60.9 | 57.2 | 59.0 |

Table 8: 10-fold cross-validation on the `AImed` data set.

Bunescu (2007) reports the evaluation results on the `AImed` corpus in the form of a precision-recall curve. If we consider the highest precision that was obtained in our experiments (69.09 or 71.16, depending on the input), this roughly corresponds to a recall of 35% in his plot (referred to as Baseline I in Table 8). In sum, the shortest path approach never approaches performance of the LA kernel on any of the biomedical data sets that were studied here. The other baseline, Baseline II, achieves the lowest scores from all the methods presented here.

Table 8 illustrates that not only various methods have been trained on the `AImed` corpus, but also many different parsers have been used. It should be noted that the graph kernel has been trained and tested on the syntactic representation generated by the Charniak-Lease parser, and the shortest path kernel has explored dependency paths obtained from the Collins parser. The Charniak-Lease parser is a statistical parser trained on the biomedical data (Lease & Charniak, 2005), whose phrase structures can be transformed into dependencies. Likewise, the Collins parser is a statistical parser (Collins, 1999). This leads to the question whether the choice of syntactic parser has a significant impact on the extraction results. To compare the impact of the syntactic parsers on relation extraction for `AImed`, Miyao et al. (2008) have conducted a complex study with eight parsers (including the Stanford analyzer) and five parse representations [8]. They consider two cases. In the first one, parsers have not been trained on biomedical data. Regardless of the parser being used in their experiments, accuracy for the extraction task is similar. In the second experiment, parsers have been re-trained on domain-specific data. In this case, it has been shown that the relation extraction results can be improved. The actual gain, however, can vary from one parser to another.

---

8. These are either various dependency tree formats (e. g., in the Stanford dependency format), or phrase structures, or predicate-arguments structures.

For the `AImed` data, the LA kernel with the *Dice* measure gives state-of-the-art results. It is outperformed only by approaches that use more information than just dependency paths.

### 5.3 LA Kernel Parameters

Saigo et al. (2004) have already shown that the scaling parameter $\beta$ (Equation 11) has a significant impact on accuracy. We have also carried out additional experiments by varying gap values and the value of $\beta$. Results are visualized in Figure 5. The opening and extension gap values are separated by the slash symbol and the values on the X-axis in the form '$a/b$' should be read as "the opening gap is set to $a$ and the extension gap is equal to $b$". The kernel matrices were normalized and all examples were weighted. According to our previous experiments, the results yielded by the *Dice* measure do not significantly differ from the ones achieved by the *Cosine* measure and we selected the *Dice* measure to conduct all experiments. The performance on the `BC-PPI` data set is shown in Figure 5.

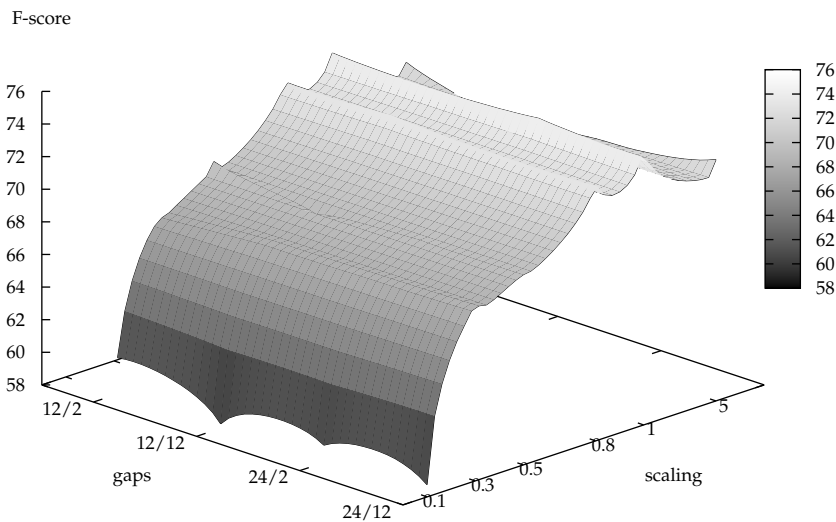

Figure 5: Varying gaps and the scaling ($\beta$) parameter on the `BC-PPI` data set (10-fold cross-validation): F-score.

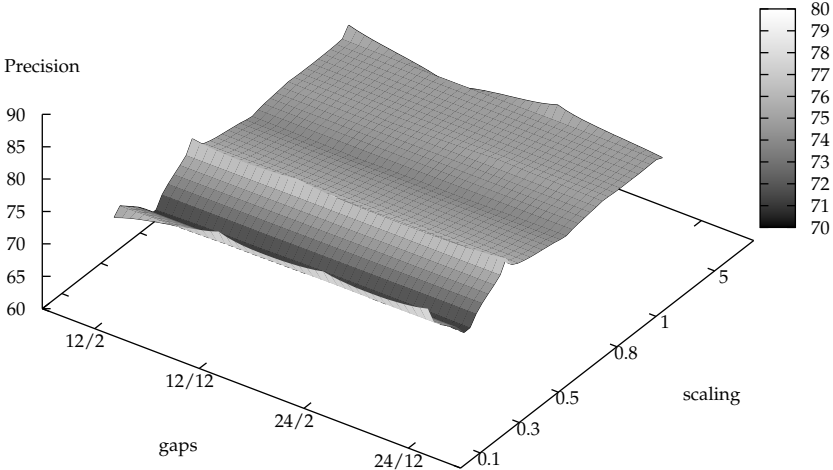

Figure 6: Varying gaps and the scaling ($\beta$) parameter on the `BC-PPI` data set (10-fold cross-validation): Precision.

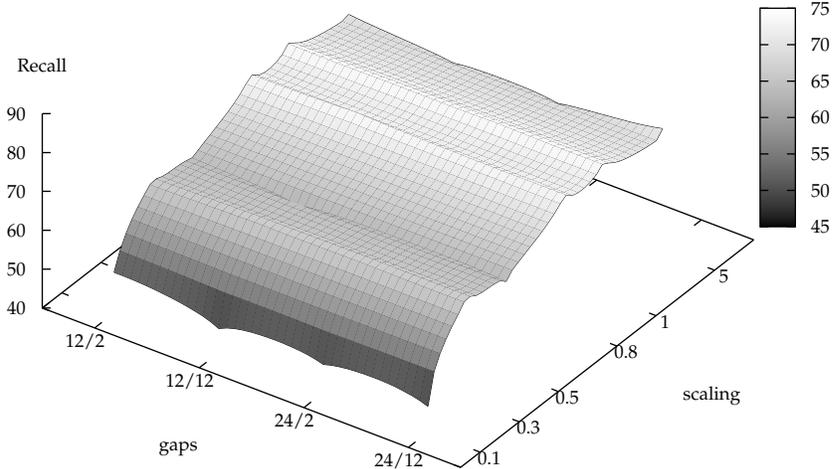

Figure 7: Varying gaps and the scaling ($\beta$) parameter on the `BC-PPI` data set (10-fold cross-validation): Recall.

The results in Figure 5 indicate that decreasing $\beta$ leads to a decrease in overall performance. Moreover, varying gap values causes subtle changes in the F-score, but these changes are not as drastic as changes due to the lower $\beta$.

Changes in the F-score are more likely to be explained by variances in precision and recall. To investigate this matter, we look at how both measures depend on parameter changes. If $\beta$ is set to a low value, one can expect that this will nearly diminish the impact of the substitution matrix, i.e. similarity among elements. For this reason we hypothesize that larger values of the scaling parameter $\beta$ should result in higher recall. Indeed, Figure 7 supports this hypothesis and the recall plot resembles the one for the F-score. Varying parameter values has a much lower impact on precision (Figure 6) but nonetheless precision does decrease as the $\beta$ parameter becomes larger.

Overall, $\beta$ seems to influence the final results the most, although gap values make a contribution as well. According to the results we obtained, setting an extension gap $e$ to a large value (or equal to the opening gap $o$) is undesirable. Since the scaling parameter $\beta$ is applied not only to the substitution matrix but to the gap values as well, setting $\beta$ below 0.5 decreases the effects of gap penalization and similarity of elements. Consequently, the best performance is achieved by setting $\beta$ to 1. This suggests that the final performance of the LA kernel is influenced by a combination of parameters and their choice is crucial for obtaining good performance.

## 6. Experiment II: Generic Relations

Another series of experiments was carried out on seven generic relations from the `SemEval - 2007` challenge, Task 4. The choice of the data sets in this case was motivated by two factors. First, semantic relations used here differ from the relations from the biomedical domain. Second, since the arguments of relations are annotated with WordNet, it becomes possible to explore information from WordNet and use it as prior knowledge for the LA kernel.

Many participants of this challenge considered WordNet either explicitly (Tribble & Fahlman, 2007; Kim & Baldwin, 2007), or as a part of a complex system (Giuliano et al., 2007). Since it is not always obvious how to use WordNet so that it yields the best performance, many researchers have made additional decisions such as use of supersenses (Hendrickx et al., 2007), selection of a predefined number of high-level concepts (Nulty, 2007), or cutting the WordNet hierarchy at a certain level (Bedmar et al., 2007). Some other systems such as the one by Nakov (2007) were based solely on information collected from the Web. Even though it became evident that the best performing systems used WordNet, the variance in the results is remarkable and it is not clear whether this difference in performance can be explained by the machine learning methods being used, the combination of features, or by some other factors.

The `SemEval-2007` Task-4 data set includes some relation examples which are nominal compounds (like 'coffee maker'), and this greatly reduces availability of information between two arguments in the dependency paths. The relation arguments in this case are linked by one grammatical relation (e.g., 'coffee' and 'maker' are linked by the grammatical relation 'nn', which corresponds to 'noun compound'). We assume, therefore, information coming from WordNet to be especially helpful when the dependency paths are that short. In all our experiments we used 5 relatedness measures defined earlier in Section 3.2 plus one additional measure which is called 'random'. The random measure indicates that the relatedness values between any two relation arguments were generated randomly (within $[0, 1]$) and is thus very suitable as a baseline (Baseline II). Similarly to the experiments in the biomedical domain, another baseline is the shortest path kernel (Baseline I). Note that in the Task 4 overview paper, Girju et al. (2007) reported on three baselines, which, in their case, were (i) guessing 'true' or 'false' for all examples, depending on which class is the majority class in the test set (Baseline III), (ii) always guessing 'true' (Baseline IV), and (iii) guessing 'true' or 'false' with the probability that corresponds to the class distribution in the test set (Baseline V).

The first question of interest is what implications the choice of semantic relatedness measure has for the performance of the LA kernel. To answer this question, we perform 10-fold cross-validation on the training set (Figure 9, Figure 10 and Figure 11). Among all 5 measures only *jcn* and *resnik* fail to perform better than the random score. In most cases, the Resnik score is outperformed by other measures. The behaviour of the Leacock-Chodorow score (*lch*) and *jcn* varies from one semantic relation to another. For instance, use of *jcn* seems to boost precision for CAUSE-EFFECT, PART-WHOLE, PRODUCT - PRODUCER, and THEME - TOOL. For the remaining three relations it is clearly not the best-performing measure.

To check whether there are differences between relatedness measures, we have carried out significance tests comparing all measures for all relations. Our findings are summarized in Table 9. Here, the symbol $\sim$ between two relatedness measures stands for the measure equivalence, or, in other words, indicates that there is no significant difference. Similarly to the experiments in the biomedical field, all significance tests were conducted using a two-tailed paired $t$-test with confidence level 95%. In addition, for any two measures $a$ and $b$, $a > b$ means that $a$ performs significantly better than $b$. For instance, the ranking for CAUSE - EFFECT in Table 9 should be read as follows. The two best performing measures are *wup* and *lch*, which significantly outperform *lin*, followed by *random* and *res*, which, in turn, yield significantly better results than *jcn*. It can be seen from this table that *wup* and *lch* are clearly the best performing measures for all seven relations (each of them is the best measure for six out of seven relations).

| Relation type | Ranking |
|---|---|
| CAUSE - EFFECT | *wup $\sim$ lch > lin > res $\sim$ random > jcn* |
| INSTRUMENT - AGENCY | *wup $\sim$ lch > lin > res > jcn $\sim$ random* |
| PRODUCT - PRODUCER | *wup $\sim$ lch > lin $\sim$ jcn $\sim$ res > random* |
| ORIGIN - ENTITY | *wup $\sim$ lch > lin > res $\sim$ jcn > random* |
| THEME - TOOL | *lch > lin $\sim$ wup > res > jcn > random* |
| PART - WHOLE | *wup $\sim$ lin $\sim$ lch > res > jcn $\sim$ random* |
| CONTENT - CONTAINER | *wup > lch > lin $\sim$ res > jcn $\sim$ random* |

Table 9: Ranking of the relatedness measures with respect to their accuracy on the training sets ($\sim$ stands for measure equivalence, $a > b$ indicates that the measure $a$ significantly outperforms $b$).

For each relation, we applied the best performing measure on the training set for this particular relation to the test data. The results are reported in Table 10. On average, the LA kernel employing the WordNet relatedness measures significantly outperforms two baselines. Moreover, when compared to the best results of the `SemEval-2007` competition (Beamer et al., 2007), our method approaches performance yielded by the best system (best$_{SV}$). This system used not only various lexical, syntactic, and semantic feature sets, but also expanded the training set by adding examples from many different sources. We have already mentioned in Section 2 that the recent work by Ó Séaghdha (2009) explores WordNet structure and graph kernels to classify semantic relations. The overall performance which is achieved by this method (Table 10) is comparable to the one by the LA kernel, but it is unclear whether there are any semantic relations for which one of the approaches performs better.

| Relation type | Accuracy | Precision | Recall | F-score | measure |
|---|---|---|---|---|---|
| Cause - Effect | 61.25 | 62.50 | 60.98 | 61.73 | lch |
| Instrument - Agency | 75.64 | 73.17 | 78.95 | 75.95 | wup |
| Product - Producer | 75.27 | 76.71 | 90.32 | 82.96 | lch |
| Origin - Entity | 74.07 | 75.86 | 61.11 | 67.69 | wup |
| Theme - Tool | 73.24 | 67.86 | 65.52 | 66.67 | lch |
| Part - Whole | 80.56 | 70.00 | 80.77 | 75.00 | wup |
| Content - Container | 71.62 | 74.29 | 68.42 | 71.23 | wup |
| Average | 73.09 | 71.48 | 72.30 | 71.60 | |
| Baseline I | 58.23 | 52.50 | 54.30 | 49.19 | |
| Baseline II | 55.83 | 61.61 | 55.50 | 53.93 | |
| Baseline III | 57.0 | 81.3 | 42.9 | 30.8 | |
| Baseline IV | 48.5 | 48.5 | 100.0 | 64.8 | |
| Baseline V | 48.5 | 48.5 | 57.1 | 48.5 | |
| best$_{SV}$ | 76.3 | 79.7 | 69.8 | 72.4 | |
| Gap-weighted string kernel (Lodhi et al., 2002) | 61.19 | 66.2 | 47.52 | 43.02 | |
| WordNet kernels (Ó Séaghdha, 2009) | 74.1 | - | - | 71.0 | |

Table 10: Results on the `SemEval-2007`, Task 4 test data set (selecting the best performing measure on the training set for each relation).

In addition, we report results on the `SemEval` Task 4 test set per relatedness measure (Table 11), which are averages over all seven relations. Similarly to our findings on the training set, *wup* and *lch* are the best performing measures on test data as well.

One would expect that the optimal use of prior knowledge should allow us to reduce the number of training instances without significant changes in performance. To study how (and whether) the amount of training data influences the results on the test set, we split the training set in several subsets, creating a model for each subset and applying it to the `SemEval-2007`, Task 4 test data. The split corresponds to the split used by the challenge organizers. As Figure 8 [9] suggests, most relations are recognized well even when a relatively small data sample is used. The exception is the Theme-Tool relation where increasing the training data clearly helps. This finding is in line with the results of Giuliano et al. (2007) whose system was a combination of kernels on the same data. Their results also indicate that all relations but one (THEME-TOOL) are extracted well, even if only a quarter of the training set is used.

---

9. The model trained on only 35 Origin-Entity examples classifies none of the test examples as positive, for this reason there is no point in Figure 8 for this relation given 35 training examples.

| Relatedness measure | Accuracy | Precision | Recall | F-score |
|---|---|---|---|---|
| wup | 72.91 | 71.20 | 72.56 | 71.62 |
| lch | 72.96 | 72.31 | 70.93 | 71.02 |
| lin | 65.27 | 62.01 | 67.07 | 63.65 |
| res | 62.94 | 62.51 | 59.66 | 60.46 |
| jcn | 55.55 | 52.25 | 69.28 | 57.07 |
| random | 56.57 | 53.10 | 52.94 | 52.83 |

Table 11: Results on the `SemEval-2007`, Task 4 test data set, averages for all 7 relations per WordNet relatedness measure.

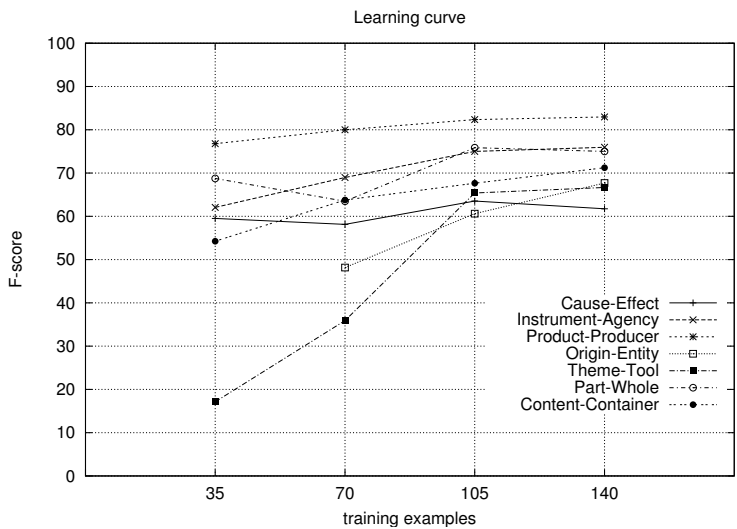

Figure 8: Learning curve on the `SemEval-2007`, Task 4 test data set.

Some other recent work on the `SemEval` Task 4 data set includes investigation of distributional kernels (Ó Séaghdha & Copestake, 2008), pattern clusters (Davidov & Rappoport, 2008), relational similarity (Nakov & Hearst, 2008), and WordNet kernels. Unlike WordNet kernels, the first three approaches do not use WordNet. Ó Séaghdha and Copestake (2008) report an accuracy of 70.7 and the F-score of 67.5 as the best results yielded by distributional kernels and the best performance of Davidov and Rappoport's (2008) method is an accuracy of 70.1, and the F-score of 70.6. WordNet kernels, similarly to our findings with the LA kernel, yield better accuracy than methods not using WordNet (74.1), but the

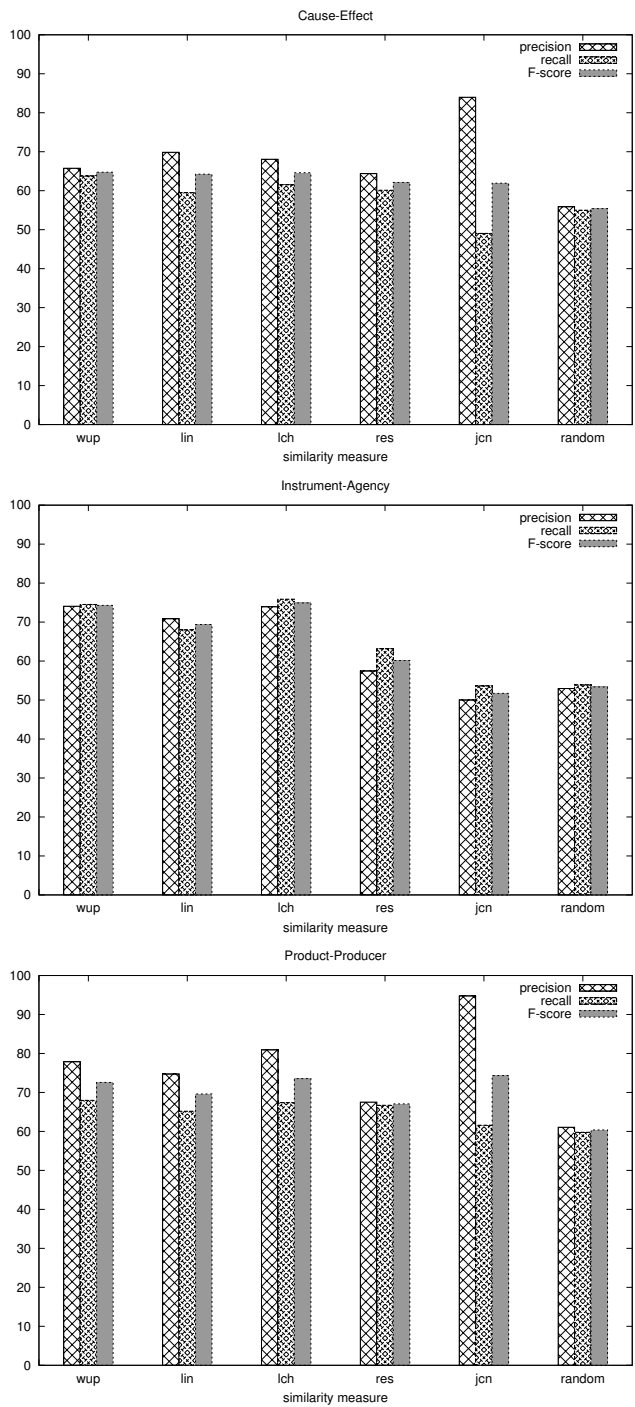

Figure 9: 10-fold cross-validation on the training set (Cause - Effect, Instrument - Agency and Product - Producer relations).

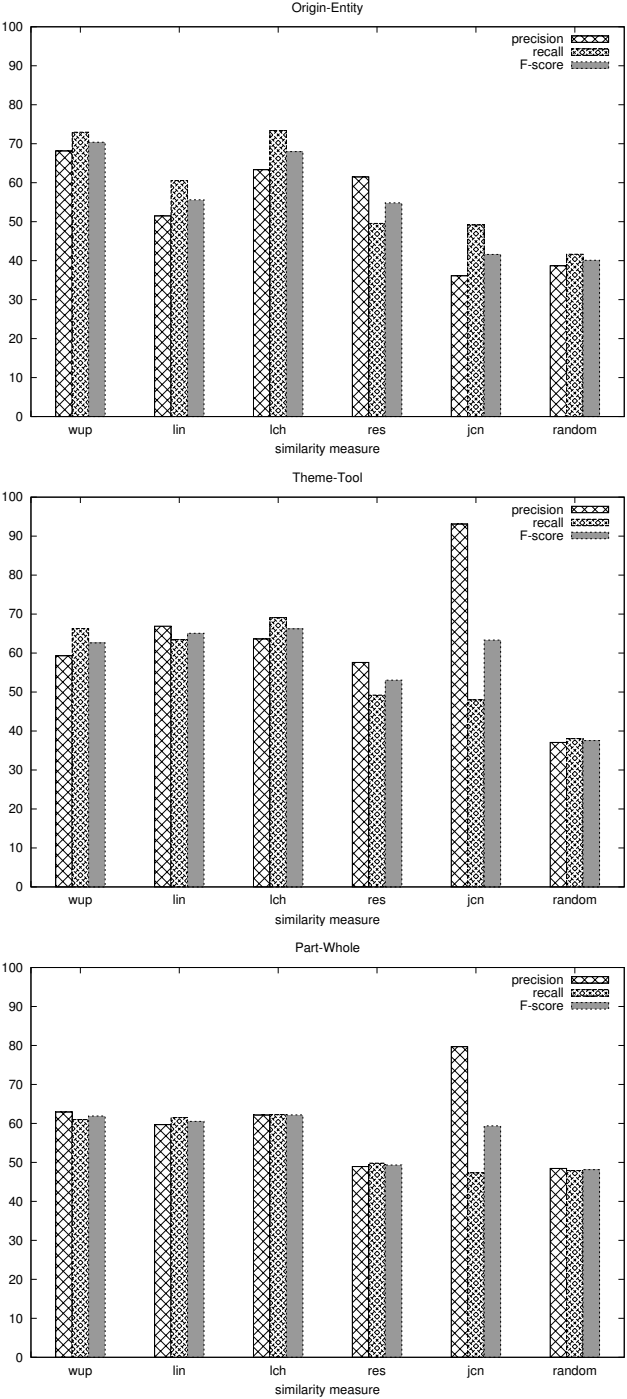

Figure 10: 10-fold cross-validation on the training set (ORIGIN - ENTITY, THEME - TOOL and PART - WHOLE relations).

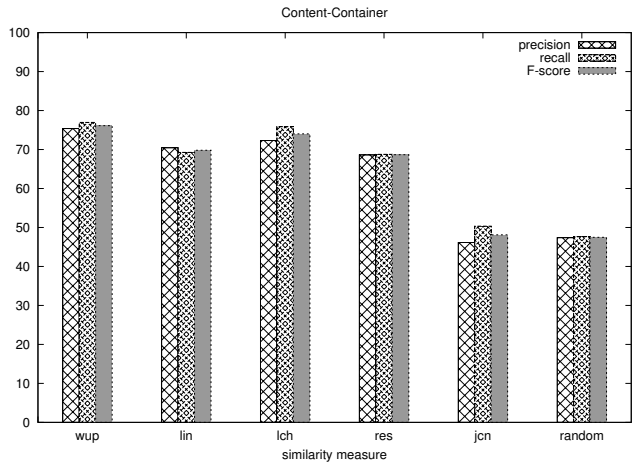

Figure 11: 10-fold cross-validation on the training set (Content - Container relation).

F-score is comparable to the performance reported by Ó Séaghdha and Copestake (2008) and by Davidov and Rappoport (2008).

## 7. Discussion

In this section we revisit the goals that were stated at the end of Section 2 and discuss our findings in more detail.

### 7.1 The LA Kernel for Relation Extraction

We have introduced the LA kernel, which has proven to be effective for biomedical problems, in the NLP domain and showed that it is well suited for relation extraction. In particular, the experiments in two different domains either outperform existing methods or yield performance on par with existing state-of-the-art kernels.

One of the motivations for using the LA kernel in the relation extraction task is to exploit prior knowledge. Here, we explore two possibilities, distributional similarity and information provided by WordNet.

#### 7.1.1 Distributional Similarity Measures

In our setting, we consider three distributional measures that have already been studied before. For instance, Lee (1999) uses them to detect similar nouns based on verb-object co-occurrence pairs. The results suggest the *Jaccard* coefficient (which is related to the *Dice* measure) to be one of the best performing measures followed by some others including *Cosine*. *Euclidean* distance fell into the group with the largest error rates. Given previous work by Lee (1999), one would expect Euclidean distance to achieve worse results than the other two measures. Indeed, on the `LLL` corpus, the LA kernel employing *L2* shows a significant decrease in performance. As to the other measures, the method using *Dice* significantly outperforms the one based on the *L2* measure only on the `LLL` corpus while there is no significant improvement on the `BC-PPI` data set. Based on the experiments we have conducted, we conclude that the LA kernel using *Dice* and *Cosine* measures performs similarly on the `LLL` data set and the `BC-PPI` corpus. Given the results on various biomedical corpora (and different settings we have experimented with), we obtained experimental support for choosing the *Dice* or *Cosine* measure over the *Euclidean* distance.

### 7.1.2 WORDNET SIMILARITY MEASURES

For generic relations, semantic relatedness plays a significant role. The difference in the F-score between models that use semantic relatedness and the kernel where the relatedness values are generated randomly (Baseline II) amounts to nearly 20%. All measures exhibit different performance on the seven generic relations that we have considered. We can observe, for instance, that *wup*, *lch*, and *lin* almost always yield the best results, no matter what relation is considered. We found the Resnik score and Jiang and Conrath's measure to yield lower results than other measures. Even though the F-scores per relation vary quite substantially (by placing CAUSE-EFFECT, THEME-TOOL, ORIGIN-ENTITY among the most difficult relations to extract), two measures, *wup* and *lch*, are the top-performing measures for all seven relations. These two measures explore the WordNet taxonomy using a length of the paths between two concepts, or their depth in the WordNet hierarchy and, consequently, belong to the path-based measures. The other three measures, *res*, *lin* and *jcn* are information content based measures, and here relatedness between two concepts is defined through the amount of information they share. Our experiments with the LA kernel on generic relation recognition suggest that, in this particular case, the path-based measures should be preferred over the information content based measures.

We should stress, however, that this is the evaluation of the semantic relatedness measures in the context of relation recognition, and one can by no means draw a conclusion that the top measures for other NLP tasks will stay the same. For example, Budanitsky and Hirst (2006) use semantic relatedness measures to detect malapropism and show that Jiang and Conrath's measure (*jcn*) yields the best results, followed by Lin's measure (*lin*), and the one by Leacock and Chodorow (*lch*), and then by Resnik's measure (*res*). Our results are quite similar to their findings if we consider the *res* measure, but *jcn* is not on the top of the accuracy ranking list for any of the seven semantic relations that we have studied.

## 7.2 Factors and Parameters that Influence the LA Kernel Performance

Our experiments in two domains have shown that the LA kernel either outperforms existing methods on the same corpora, or yields performance on par with existing state-of-the-art kernels.

### 7.2.1 BASELINES

An advantage of the LA kernel over the Bunescu shortest path method (Baseline I) is that it is capable of handling paths of different lengths. By allowing gaps and penalizing them, the final kernel matrix becomes less sparse. The shortest path approach also attempts to generalize over the dependency paths, but it usually overgeneralizes which leads to high recall scores (Table 5 and Table 6) but to poor overall performance. One explanation for overgeneralization may be that this method accounts well for structural similarity (provided sequences are of the same length) but fails to provide finer distinctions among dependency paths. Consider, for example, two sequences 'trip ← makes → tram' and 'coffee ← makes → guy', whereby the first path represents a negative instance of the PRODUCT-PRODUCER relation and the second path corresponds to a positive one. Even though they do not match exactly, the elements that do not match are all nouns in singular. Consequently, comparison according to the shortest path method will result in a relatively high similarity score. In contrast, the LA kernel will consider similarity of the elements and the pairs 'trip'-'coffee' and 'tram'-'guy' will obtain low scores.

In addition, Baseline II, which is based on randomly generated substitution scores, performs poor for all data sets (or comparable to Baseline I). This leads us to the conclusion that accurate estimation of similarities is another reason why the LA kernel performs well on relation extraction.

### 7.2.2 COMPARISON WITH OTHER METHODS

As we have already pointed out, the obvious shortcoming of Baseline I is its inability to handle dependency paths of different length. For this reason, we have also applied the gap-weighted string kernel (Lodhi et al., 2002) to all data sets. In this case, dependency paths can be compared in a flexible way because gapping is allowed, but no other additional information is used. This kernel outperforms Baseline I by increasing precision of relation extraction while preserving a relatively high recall. The only data set where it fails to yield good results is the `LLL` test data, and we believe this is due to the differences in the `LLL` training and test data. For all data sets, the LA kernel achieves better performance than the gap-weighted string kernel. The margin, however, is different for different data sets. In the biomedical domain, the differences between the two methods can more clearly be seen on the `BC-PPI` and `LLL` data sets, while the results on the `AImed` corpus are comparable. However, other methods tested on `AImed` do not get higher scores unless they use more features than just dependency paths. This holds for both types of cross-validation used on this corpus. For generic relations, the difference between the LA kernel and the gap-weighted string kernel is much larger. In particular, in the case of the gap-weighted kernel, precision is high, but recall is much lower. This can be explained by the fact that generic relations benefit from the knowledge found in WordNet and recall achieved by the LA kernel is, therefore, high. The gap-weighted kernel has access only to information found in the dependency paths and, for this reason, fails to find more relations.

The LA kernel also achieves the best performance on the `LLL` training set, outperforming the graph kernel (Airola et al., 2008), the shallow linguistic kernel (Giuliano et al., 2006) and the rule-based system by Fundel et al. (2007). All three have used different input for their methods, varying from plain text to dependency structures. For this reason, a direct comparison is unfortunately not possible, but we can conclude that the methods employing dependency information always are among the best performing approaches.

Two other approaches whose performance has been reported on the `AImed` data set include the tree kernel (Sætre et al., 2008) and TSVM (Erkan et al., 2007). Both of them explore syntactic information in different ways. While Sætre et al. consider subtrees, the method of Erkan et al. has more similarities with our approach because it relies on the dependency path comparison. To do this comparison, they only use information already available in the dependency paths (SVM setting), or more dependency paths (TSVM setting). According to Lauer and Bloch (2008), TSVMs fall into the category using prior knowledge by 'sampling methods', because it explores prior knowledge by generating new examples. In contrast, we employ information from large unlabeled text sources in order to enable finer comparison of the dependency paths and always work in the supervised learning setting. Using the same evaluation procedure as in the work of Sætre et al. and Erkan et al. we show that the LA kernel outperforms both methods, but the differences on this data set are much smaller than on the other data sets we have used.

### 7.2.3 The LA Parameters

We have demonstrated that the choice of LA parameters is crucial for achieving good performance. In our experiments, the scaling parameter $\beta$ contributes to the overall performance at most, but the other parameters such as gap values have to be taken into account as well. When $\beta$ approaches infinity, the LA kernel approximates the Smith-Waterman distance, but increasing $\beta$ does not necessarily have a positive impact on the final performance. This finding is in line with the results reported by Saigo et al. (2004) on the homology detection task. The best performance is yielded by setting the scaling parameter to 1 or a bit higher, and by penalizing the gap extension less than the gap opening.

## 8. Conclusions and Future Work

We have presented a novel approach to relation extraction that is based on the local alignments of sequences. Using an LA kernel provides us an opportunity to explore various sources of information and to study their role in relation recognition. Possible future directions include, therefore, an examination of other distributional similarity measures, studying their impact on the extraction of generic relations, and looking for other sources of information which could be helpful for relation recognition. It may be interesting to consider relational similarity (Turney, 2006), which looks for the correspondence between relation instances. In this case, one should be able to infer that 'doctor' corresponds to 'scalpel' in a similar way as 'fisherman' to 'net' (where both (scalpel, doctor) and (net, fisherman) are examples of Instrument - Agency).

Despite the sparseness problem that might occur when WordNet-based measures are used, these measures have an advantage over the distributional measures by treating elements to be compared as concepts rather than words. In the NLP community, a few steps have been already taken to solve this problem by clustering words in large corpora aiming at word sense discovery (Pennacchiotti & Pantel, 2006). Recently, Mohammad (2008) in his thesis investigated the compatibility of distributional measures with ontological ones. By using corpus statistics and a thesaurus, the author introduced distributional profiles of senses and defined distance measures on them. Even though this new approach to calculat-ing similarity was tested on generic corpora, it would be of a certain interest to apply it to domain-specific data.

Overall, local alignment kernels provide a flexible means to work with data sequences. First, they allow a partial match between sequences which is particularly important when dealing with text. Second, it is possible to incorporate prior knowledge in the learning process while preserving kernel validity. In general, LA kernels can be applied to other NLP problems as long as the input data is in the form of sequences.

## Acknowledgments

The authors wish to thank Simon Carter and Gerben de Vries for their comments and proofreading, and three anonymous reviewers for their highly valuable feedback. They also acknowledge the input from the Adaptive Information Management (AIM) group at the University of Amsterdam. The preliminary version of this work has been dicussed at the 22nd International Conference on Computational Linguistics (CoLing 2008) and at the Seventh International Tbilisi Symposium on Language, Logic and Computation (2007). This work was carried out in the context of the Virtual Laboratory for e-Science project (`www.vl-e.nl`). This project is supported by a BSIK grant from the Dutch Ministry of Education, Culture and Science (OC&W) and is part of the ICT innovation program of the Ministry of Economic Affairs (EZ).